\documentclass[]{fairmeta} % For LaTeX2e

\usepackage{amsmath,amsfonts,bm}

\def\eqref#1{equation~\ref{#1}}
\def\1{\bm{1}}

\DeclareMathAlphabet{\mathsfit}{\encodingdefault}{\sfdefault}{m}{sl}
\SetMathAlphabet{\mathsfit}{bold}{\encodingdefault}{\sfdefault}{bx}{n}

\usepackage[utf8]{inputenc}  % allow utf-8 input
\usepackage[T1]{fontenc}     % use 8-bit T1 fonts
\usepackage{hyperref}        % hyperlinks
\usepackage{url}             % simple URL typesetting
\usepackage{booktabs}        % professional-quality tables
\usepackage{amsfonts}        % blackboard math symbols
\usepackage{nicefrac}        % compact symbols for 1/2, etc.
\usepackage{microtype}       % microtypography
\usepackage{graphicx}
\usepackage{multicol}
\usepackage{multirow}
\usepackage{xspace}
\usepackage{wrapfig}
\usepackage{enumitem}
\usepackage[dvipsnames,table]{xcolor}
\usepackage{colortbl}
\usepackage{caption}
\usepackage{lmodern} 
\usepackage{algorithm}
\usepackage{algorithmic}
\usepackage{subcaption}
\usepackage{needspace}
\usepackage{wrapfig}
\usepackage{xspace}

\usepackage{multicol}
\usepackage{multirow}
\usepackage{pifont}
\usepackage{tabularx}
\usepackage{makecell}
\usepackage{array}
\newcolumntype{R}{>{\raggedleft\arraybackslash}X}
\definecolor{lightblue}{RGB}{220,235,250}
\definecolor{lightgray}{gray}{0.91}

\newcommand{\cmark}{{\textcolor{green!70!black}{\ding{51}}}}
\newcommand{\pmark}{{\textcolor{olive}{\ding{51}}}}
\newcommand{\xmark}{{\textcolor{red}{\ding{55}}}}
\usepackage[most,skins,theorems]{tcolorbox}
\usepackage{listings}
\newtcolorbox{promptbox}[1][]{
  enhanced, breakable,
  colback=gray!1,      
  colframe=gray!60,    
  coltitle=black,      
  boxrule=2pt,
  arc=10pt,
  left=6pt, right=6pt, top=6pt, bottom=6pt,
  title={#1}, fonttitle=\bfseries,
  attach boxed title to top left={yshift*=-3mm},
  boxed title style={colback=gray!10}
}

\newtcolorbox{templatebox}[1]{
  enhanced,
  breakable,
  colback=white,
  colframe=black!65,
  colbacktitle=black!80,
  coltitle=white,
  boxrule=0.9pt,
  arc=2pt,
  left=6pt,
  right=6pt,
  top=6pt,
  bottom=6pt,
  title={#1},
  fonttitle=\bfseries,
  sharp corners,
  boxed title style={sharp corners, boxrule=0pt}
}

\newcommand{\promptsection}[1]{\par\smallskip\noindent\textbf{\# #1}\par}
\newlength{\loglabelwidth}
\newcommand{\logentry}[2]{%
  \Needspace{4\baselineskip}%
  \par\noindent
  \makebox[\loglabelwidth][l]{\ttfamily\scriptsize #1}%
  \hangindent=\dimexpr\loglabelwidth+0.7em\relax
  \hangafter=1
  \hspace{0.7em}#2\par
}
\newenvironment{logblock}{%
  \par\smallskip
  \begingroup
  \small
  \setlength{\parindent}{0pt}
  \setlength{\parskip}{2pt}
}{%
  \par
  \endgroup
}

\lstdefinestyle{promptplain}{
  language={},
  basicstyle=\ttfamily\footnotesize,
  keywordstyle=\color{black},
  commentstyle=\color{gray},
  stringstyle=\color{black},
  backgroundcolor=\color{gray!5},
  frame=single,
  rulecolor=\color{black!35},
  numbers=none,
  breaklines=true,
  showstringspaces=false
}

\tcbset{
  aibox/.style={
    width=\linewidth,
    top=8pt,
    bottom=4pt,
    colback=inftythink-red!15,
    colframe=inftythink-red,
    colbacktitle=inftythink-red!90!black,
    enhanced,
    center,
    attach boxed title to top left={yshift=-0.1in,xshift=0.15in},
    boxed title style={boxrule=0pt,colframe=white,},
  }
}
\newtcolorbox{AIbox}[2][]{aibox,title=#2,#1}
\usepackage{amsthm}
\theoremstyle{plain}

\theoremstyle{definition}

\theoremstyle{remark}

\newcommand{\methodname}{KnowU-Bench\xspace}

\title{
{\methodname}: Towards Interactive, Proactive, and Personalized Mobile Agent Evaluation
}

\author{Tongbo Chen$^{1*}$,
Zhengxi Lu$^{1*}$,
Zhan Xu$^{1*}$,
Guocheng Shao$^{1*}$,
Shaohan Zhao$^{1*}$,\\
Fei Tang$^{1}$,
Yong Du$^{1}$,
Kaitao Song$^{2}$,
Yizhou Liu$^{1}$,
Yuchen Yan$^{1}$,
Wenqi Zhang$^{1}$, \\
Xu Tan$^{3}$,
Weiming Lu$^{1}$,
Jun Xiao$^{1}$,
Yueting Zhuang$^{1}$,
Yongliang Shen$^{1\dagger}$}

\affiliation[1]{Zhejiang University}
\affiliation[2]{Apple}
\affiliation[3]{Tencent}

\contribution[*]{Equal Contribution}
\contribution[\dagger]{Corresponding authors}

\abstract{

Personalized mobile agents that infer user preferences and calibrate proactive assistance hold great promise as everyday digital assistants, yet existing benchmarks fail to capture what this requires. 
Prior work evaluates preference recovery from static histories or intent prediction from fixed contexts. Neither tests whether an agent can elicit missing preferences through interaction, nor whether it can decide when to intervene, seek consent, or remain silent in a live GUI environment.
We introduce \textbf{\methodname}, an online benchmark for personalized mobile agents built on a reproducible Android emulation environment, covering 42 general GUI tasks, 86 personalized tasks, and 64 proactive tasks. Unlike prior work that treats user preferences as static context, \methodname hides the user profile from the agent and exposes only behavioral logs, forcing genuine preference inference rather than context lookup. To support multi-turn preference elicitation, it instantiates an LLM-driven user simulator grounded in structured profiles, enabling realistic clarification dialogues and proactive consent handling. Beyond personalization, \methodname provides comprehensive evaluation of the complete proactive decision chain, including grounded GUI execution, consent negotiation, and post-rejection restraint, evaluated through a hybrid protocol combining rule-based verification with LLM-as-a-Judge scoring.
Our experiments reveal a striking degradation: agents that excel at explicit task execution fall below 50\% under vague instructions requiring user preference inference or intervention calibration, even for frontier models like Claude Sonnet 4.6. The core bottlenecks are not GUI navigation but preference acquisition and intervention calibration, exposing a fundamental gap between competent interface operation and trustworthy personal assistance.

}

\date{\today}
\metadata[Page]{\url{https://zju-real.github.io/KnowU-Bench}}
\metadata[Code]{\url{https://github.com/ZJU-REAL/KnowU-Bench}}
\correspondence{\email{\{zhengxilu, syl\}@zju.edu.cn}}

\begin{document}

\maketitle

\begin{figure}[h]
\centering
\includegraphics[width=1.0\textwidth]{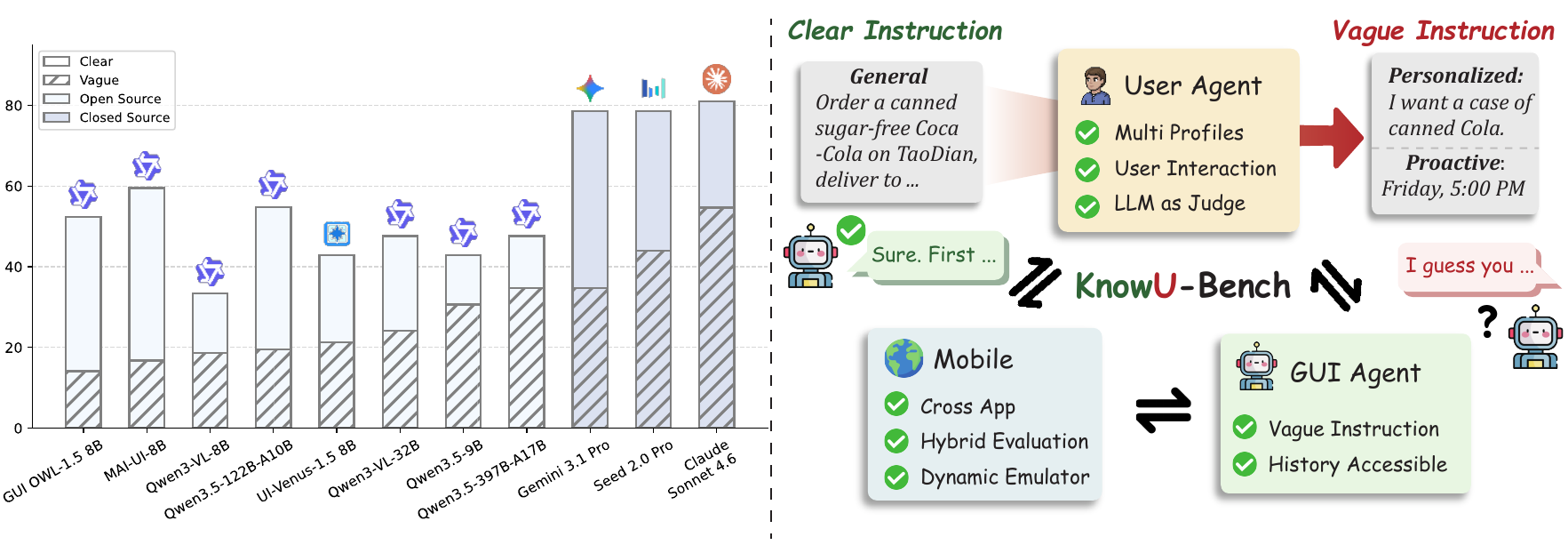}
\caption[Benchmark comparison]{\textbf{Left:} Model performance drops substantially from clear to vague instructions. \textbf{Right:} Key components of \methodname.}
\label{fig:comparison}
\end{figure}

\section{Introduction}
\label{sec:intro}

GUI agents can now navigate complex multi-step workflows, coordinate actions across multiple apps, and complete real-world tasks on mobile devices with increasing reliability~\citep{ye2025mobileagentv3,liu2025guisurvey,tang2025guisurvey,lu2025uis1,gu2025uivenus}.
Benchmarks such as AndroidWorld~\citep{rawles2024androidworld} and MobileWorld~\citep{kong2025mobileworld} have driven rapid progress along this axis, and today's strongest agents can reliably complete well-defined tasks across a broad range of real applications. 
Yet the demands of practical deployment have moved well beyond instruction following.
Products like Doubao Mobile Assistant and OpenClaw~\citep{openclaw2026} are increasingly positioned as \textit{personal} assistants that are expected to know your preferred delivery platform without being told, remember you cannot tolerate spicy food when ordering lunch, and silence your alarm on Friday nights because they have learned your weekend routine.
The question is no longer \textit{can the agent follow instructions}, but \textit{can the agent act on your behalf as if it truly understands you}.

This shift exposes a fundamental mismatch between what current benchmarks measure and what real deployment demands. 
An instruction as natural as ``order me lunch'' requires an agent to jointly resolve app preference, dietary constraints, budget, and payment habit from user history, with no explicit signal separating the right answer from a plausible but wrong one. The difficulty in proactive settings, where the agent must decide whether to act without any instruction at all.
Our experiments reveal a substantial performance gap between clear and vague instructions: as shown in the left panel of Figure~\ref{fig:comparison}, models that perform well on specified tasks degrade sharply on ambiguous, preference involved requests and proactive decisions.

% Recent efforts have begun to address personalized evaluation for mobile agents, broadly along two lines.
% The first line focuses on \textbf{xxx}...
% The second line targets \textbf{xxx}...
% While each of these efforts advances its respective direction, three systemic gaps remain unresolved across the field.

Recent efforts have begun to address personalized evaluation for mobile agents, broadly along two lines. The first line focuses on preference modeling from historical records: FingerTip 20K~\citep{yang2025fingertip20k} mines proactive task suggestions and personalized execution signals from long-term mobile usage logs, while PersonalAlign~\citep{lyu2026personalalign} and Me-Agent~\citep{wang2026meagent} treat personalization as a problem of recovering user intent from static behavioral histories. The second line targets proactive intent inference: ProactiveMobile~\citep{kong2026proactivemobile} emphasizes context-aware action prediction, and PIRA-Bench~\citep{chai2026pirabench} centers on proactive intent recommendation, with evaluation defined primarily at the level of function-sequence prediction or suggestion ranking. While each of these efforts advances its respective direction, three systemic gaps remain unresolved across the field.

\begin{enumerate}[leftmargin=*]
    \item \textbf{Personalization remains mostly offline.} Existing benchmarks focus on trajectory matching or intent similarity, rather than whether an agent completes the task correctly in a live GUI environment. The few online benchmarks are more realistic but less reproducible.
    \item \textbf{Interactive preference acquisition is not evaluated.} Existing benchmarks evaluate whether an agent can recover user intent from a static log. In practice, agents are expected to acquire missing user preferences through interaction; yet no existing benchmark evaluates this capability directly.
    \item \textbf{Proactive task remains incomplete.} Proactive task requires not only intent prediction but also calibrated initiative. Existing work still falls short of evaluating the full decision chain: whether to intervene, seek consent, or remain silent when no routine applies or the user has declined.

\end{enumerate}

We introduce \textbf{\methodname}, an online, interactive personalization benchmark for mobile agents built on a reproducible Android emulation environment. {\methodname} is grounded in three design principles that directly address the limitations above, with the right panel of Figure~\ref{fig:comparison} summarizing its key distinctions from existing personalization benchmarks.
First, every task runs in a containerized, rooted Android emulator and is verified programmatically, ensuring evaluation reflects actual GUI outcomes.
Second, an LLM driven user simulator grounded in structured user profiles provides online interactive feedback. 
Third, evaluation covers the full proactive decision chain, including grounded execution, consent handling, and post-rejection restraint.
Table~\ref{tab:comparison} provides a more detailed comparison.

\methodname comprises 42 general tasks, 86 personalized tasks, and 64 proactive tasks. As shown in the left panel of Figure~\ref{fig:comparison}, current models perform strongly on clear instructions but degrade sharply once success depends on resolving vague, preference-conditioned requests, motivating our focus on personalization and proactive assistance.
Our systematic evaluation of 11 representative models reveals three key findings: (1) General GUI execution is no longer the primary bottleneck: strong models perform well on clearly specified tasks, but drop by about 30\% on average once success depends on personalization or proactivity. (2) Personalized failures stem mainly from weak preference acquisition, with 93.8\% of Claude Sonnet 4.6 errors being clarification or partial preference failures—models struggle to ask the right questions or translate user feedback into preference aware decisions. (3) Proactive failures stem mainly from poor intervention calibration: for Claude Sonnet 4.6, 80.0\% of failures are intervention or passivity errors.

Our main contributions are summarized as follows:
\begin{itemize}[leftmargin=*]
    \item We propose \textbf{\methodname}, a mobile agent evaluation framework that tightly couples personalized reasoning with a programmatically verifiable Android emulator, providing a reproducible execution environment together with deterministic state verification.
 
    \item We construct evaluation scenarios for \emph{interactive preference acquisition} and a \emph{full proactive service decision chain}---covering unsolicited proposals, optional confirmation, grounded execution, and appropriate restraint after user rejection or in the absence of an established routine.
 
    \item We systematically evaluate 11 mainstream models on \methodname, revealing that they struggle to elicit user preferences through interaction on personalized tasks, and to calibrate when to intervene versus remain silent on proactive ones.
\end{itemize}

\section{Related Work}
% !TEX root = ../main.tex

% ── Preamble additions ────────────────────────────────────────────────────────
% \usepackage{booktabs}
% \usepackage{multirow}
% \usepackage{makecell}
% \usepackage{colortbl}
% \usepackage{xcolor}
% \usepackage{pifont}
% \usepackage{array}

% ── Color definitions ─────────────────────────────────────────────────────────
% \definecolor{guibg}{RGB}{255,255,255}        % white for GUI rows
\definecolor{guibg}{RGB}{248,251,255} 
% \definecolor{guibg}{RGB}{235,244,255}
% \definecolor{persbg}{RGB}{235,244,255}     % very light blue for personalization rows
\definecolor{persbg}{RGB}{248,251,255}
\definecolor{oursbg}{RGB}{219,234,254}       % blue highlight for Ours row
\definecolor{oursname}{RGB}{29,78,216}       % blue for Ours benchmark name
\definecolor{sectcolor}{RGB}{107,114,128}    % muted gray for section label text
\definecolor{cmarkcolor}{RGB}{22,163,74}     % green
\definecolor{pmarkcolor}{RGB}{126,126,126}    % olive
\definecolor{xmarkcolor}{RGB}{220,38,38}     % red
\definecolor{oursrule}{RGB}{37,99,235}       % blue left rule on Ours row

% ── Symbol redefinitions ──────────────────────────────────────────────────────
\renewcommand{\cmark}{\textcolor{cmarkcolor}{\ding{51}}}
\renewcommand{\pmark}{\textcolor{pmarkcolor}{\ding{51}}}
\renewcommand{\xmark}{\textcolor{xmarkcolor}{\ding{55}}}

% ── Section heading macro (full-width rule) ───────────────────────────────────
% \secrow{N}{label} — spans N columns, dash + italic label + leaders rule

% ── Section heading macro ─────────────────────────────────────────────────────
% Usage: \secrow{N}{label}
\newcommand{\secrow}[2]{%
  \midrule
  \multicolumn{#1}{l}{\small\itshape #2} \\
  \midrule
}

\begin{table}[t]
\centering
\setlength{\tabcolsep}{5pt}
\renewcommand{\arraystretch}{1.22}
\caption{%
  Comparison of \textbf{KnowU-Bench} with existing GUI benchmarks and datasets.
  \cmark~fully incorporated;\enskip
  \pmark~partially incorporated;\enskip
  \xmark~not incorporated.%
}
\label{tab:comparison}
\resizebox{\textwidth}{!}{%
\begin{tabular}{l ccccc c c}

\toprule[1.3pt]

\multirow{2}{*}{\textbf{Benchmark or Dataset}}
  & \multicolumn{5}{c}{\textbf{Capability Dimensions}}
  & \multirow{2}{*}{\makecell{\textbf{Evaluation}\\\textbf{Method}}}
  & \multirow{2}{*}{\makecell{\textbf{Task}\\\textbf{Target}}} \\

\cmidrule(lr){2-6}

  & \makecell{\textbf{Vague}\\\textbf{Instr.}}
  & \makecell{\textbf{Proactive}\\\textbf{Exec.}}
  & \makecell{\textbf{User}\\\textbf{Sim.}}
  & \makecell{\textbf{User}\\\textbf{Logs}}
  & \makecell{\textbf{User}\\\textbf{Model.}}
  & & \\

% \midrule

% ── GUI Execution ─────────────────────────────────────────────────────────────
\secrow{8}{GUI Execution Benchmarks}

\rowcolor{guibg}
AITW~\citep{rawles2023aitw}
  & \xmark & \xmark & \xmark & \xmark & \xmark
  & Action Matching & GUI Execution \\

\rowcolor{guibg}
AndroidControl~\citep{li2024androidcontrol}
  & \xmark & \xmark & \xmark & \xmark & \xmark
  & Action Matching & GUI Execution \\

\rowcolor{guibg}
SPA-Bench~\citep{chen2024spabench}
  & \xmark & \xmark & \xmark & \xmark & \xmark
  & LLM as Judge & GUI Execution \\

\rowcolor{guibg}
AndroidWorld~\citep{rawles2024androidworld}
  & \xmark & \xmark & \xmark & \xmark & \xmark
  & Rule-based & GUI Execution \\

\rowcolor{guibg}
AndroidLab~\citep{xu2025androidlab}
  & \xmark & \xmark & \xmark & \xmark & \xmark
  & Rule-based + LLM as Judge & GUI Execution \\

\rowcolor{guibg}
AndroidDaily~\citep{yan2025androiddaily}
  & \xmark & \xmark & \xmark & \xmark & \xmark
  & Action Matching + Rule-based & GUI Execution \\

\rowcolor{guibg}
MobileWorld~\citep{kong2025mobileworld}
  & \cmark & \xmark & \cmark & \xmark & \xmark
  & Rule-based & GUI Execution \\

\addlinespace[2pt]

% ── Personalization & Proactive ───────────────────────────────────────────────
\secrow{8}{Personalization \& Proactive Benchmarks}

\rowcolor{persbg}
PersonalAlign~\citep{lyu2026personalalign}
  & \cmark & \pmark & \xmark & \pmark & \pmark
  & Action Matching + LLM as Judge & Intent Alignment \\

\rowcolor{persbg}
Me-Agent~\citep{wang2026meagent}
  & \pmark & \xmark & \xmark & \pmark & \pmark
  & Action Matching & Preference Alignment \\

\rowcolor{persbg}
ProactiveMobile~\citep{kong2026proactivemobile}
  & \xmark & \pmark & \xmark & \xmark & \xmark
  & LLM as Judge & Action Prediction \\

\rowcolor{persbg}
PIRA-Bench~\citep{chai2026pirabench}
  & \xmark & \pmark & \xmark & \xmark & \xmark
  & LLM as Judge & Intent Recommendation \\

\rowcolor{persbg}
Pare~\citep{nathani2026proactive}
  & \xmark & \cmark & \cmark & \xmark & \xmark
  & Rule-based & \makecell{Proactive Interaction} \\

\rowcolor{persbg}
FingerTip~\citep{yang2025fingertip20k}
  & \xmark & \pmark & \xmark & \cmark & \xmark
  & Action Matching + LLM as Judge & Behavior Prediction \\

\midrule[0.8pt]

% ── Ours ──────────────────────────────────────────────────────────────────────
\rowcolor{oursbg}
\llap{\textcolor{oursrule}{\rule[-6.5pt]{2pt}{18pt}}\hspace{4pt}}%
\textcolor{oursname}{\textbf{KnowU-Bench (Ours)}}
  & \cmark & \cmark & \cmark & \cmark & \cmark
  & Rule-based + LLM as Judge
  & \makecell{Personalized \& Proactive\\GUI Execution} \\

\bottomrule[1.3pt]

\end{tabular}
}%
\end{table}

\subsection{Mobile Agent Benchmarks}

The evaluation of mobile GUI agents has advanced rapidly alongside the development of multimodal foundation models~\citep{qin2025uitars,lu2026uir1,tang2025guig2,wu2026gem}. Early benchmarks such as AITW~\citep{rawles2023aitw} and AndroidControl~\citep{li2024androidcontrol} established action-matching protocols for offline trajectory evaluation, providing large-scale supervision signal but limited coverage of task-level success. AndroidWorld~\citep{rawles2024androidworld} marked a significant step forward by introducing a reproducible full-stack Android environment with programmatic reward functions, enabling reliable end-to-end evaluation across real applications. Subsequent work has expanded coverage and realism: AndroidLab~\citep{xu2025androidlab} unifies evaluation across both LLM-based and multimodal agents; SPA-Bench~\citep{chen2024spabench} broadens scope to bilingual, single-app, and cross-app tasks; AndroidDaily~\citep{yan2025androiddaily} targets high-frequency daily-use scenarios; and MobileWorld~\citep{kong2025mobileworld} introduces agent-user interaction under ambiguous instructions, moving closer to real deployment conditions. More recently, MemGUI-Bench~\citep{liu2026memgui} incorporates long-term memory into mobile evaluation. Despite this progress, these benchmarks share a common limitation: tasks are formulated as one-shot, explicitly specified goals, and evaluation measures execution ability in isolation from the user-specific reasoning that practical deployment demands.

\subsection{Personalized and Proactive Benchmarks}

A separate line of work directly targets personalization and proactivity, though from angles that differ from \methodname. 
On the personalization side, PersonalAlign~\citep{lyu2026personalalign} and Me-Agent~\citep{wang2026meagent} study how agents can resolve ambiguous instructions by recovering user intent from historical preference signals, treating personalization as a static inference problem given a fixed behavioral record. 
FingerTip 20K~\citep{yang2025fingertip20k} takes a complementary view, mining long-term mobile usage logs to study proactive task suggestion alongside personalized execution. 
On the proactive side, ProactiveMobile~\citep{kong2026proactivemobile} frames context-aware intervention as an action prediction problem, while PIRA-Bench~\citep{chai2026pirabench} and Pare~\citep{nathani2026proactive} focus on intent recommendation and proactive API-level execution respectively. These efforts collectively advance preference modeling and proactive intent understanding, but they remain limited in three respects. First, evaluation is conducted offline or under constrained protocols, without verifiable grounded execution in a dynamic GUI environment. Second, none of them evaluate whether an agent can \emph{acquire} missing preferences through multi-turn clarification during task execution, as opposed to inferring them from a static log. Third, proactive assessment stops at intent prediction or suggestion ranking, leaving the full decision chain, whether to intervene, whether to seek consent, and whether to refrain after rejection, unmeasured. \methodname is designed to address all three gaps within a single, reproducible online evaluation framework.

\section{\methodname}
\label{sec:bench}
\subsection{Environment Setup}
We formulate mobile automation as a Partially Observable Markov Decision Process (POMDP) $(S,O,A,T,R)$, where $S$ is the environment state, $O$ includes the instruction and interface observations (e.g., screenshots), $A$ is the space of mobile UI actions, with the detailed action space summarized in Table~\ref{tab:action_space} of Appendix~\ref{app:action_space}. The transition function at time T is $T:S\times A\to S$, and $R:S\times A\to\{0,1\}$ indicates task completion.

\paragraph{Online Mobile emulator}
\methodname runs in a containerized Android stack built around a rooted Pixel 8 AVD and a FastAPI orchestration server. A unified controller maps agent actions to executable ADB operations and supports the full task lifecycle, from initialization to evaluation. To ensure reproducibility, each task starts from a fixed emulator snapshot and resets transient states such as backend processes, callback files, and interaction history. Time sensitive tasks additionally override device time during initialization.
\paragraph{App Coverage}
Compared with MobileWorld, \methodname expands the app ecosystem to 23 applications in total, providing broader coverage for personalized decision making, particularly in commerce and daily service scenarios. Beyond the original MobileWorld setting, we introduce one additional shopping app (\texttt{jingdian}) and two food delivery apps (\texttt{chilemei} and \texttt{tuantuan}), enabling cross-platform preference following. Detailed app information is provided in Appendix~\ref{app:information}.

\subsection{User Agent}
\begin{figure*}[t]
\centering
\includegraphics[width=\textwidth]{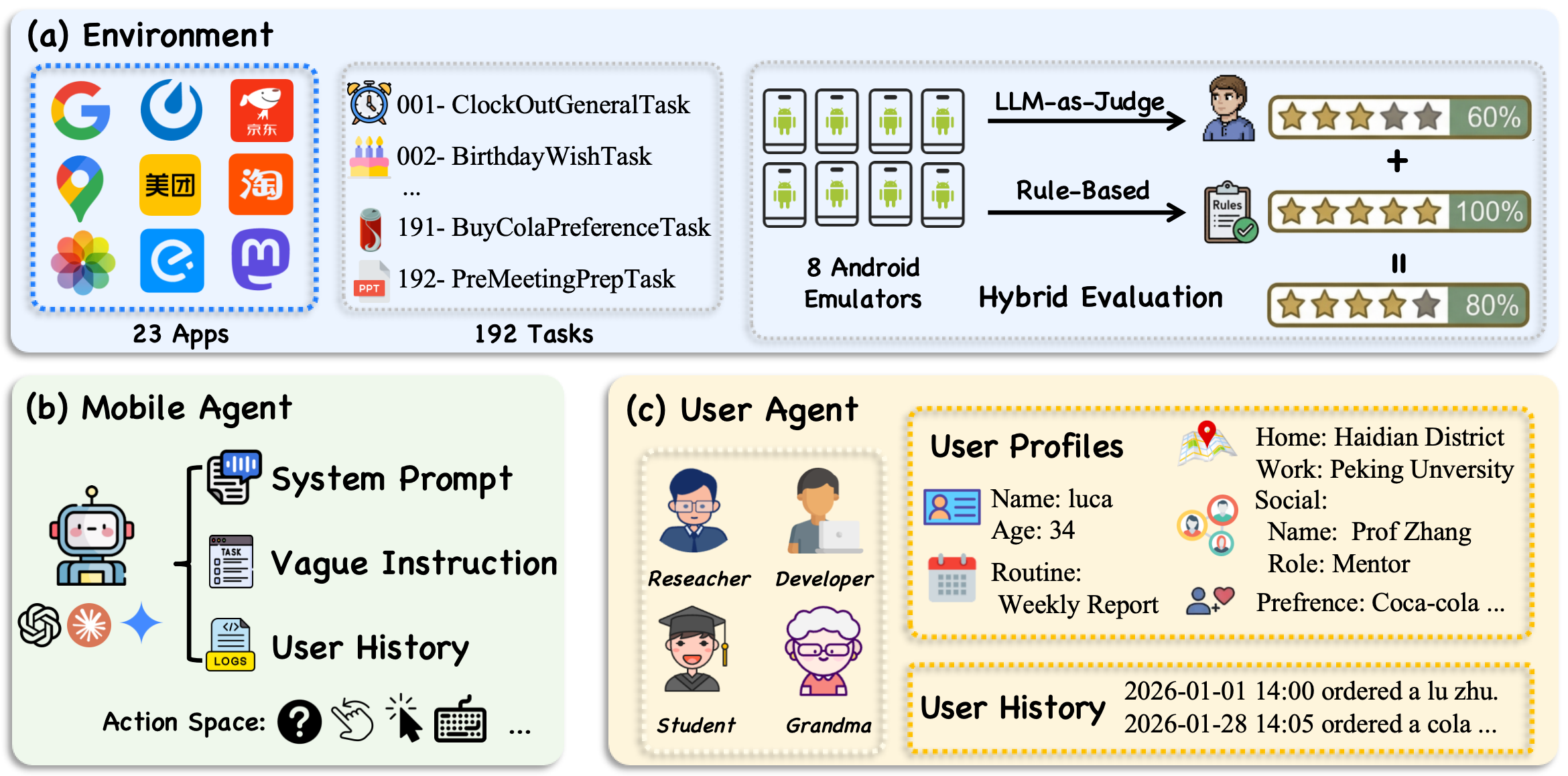}
\caption[Overview of the framework]{\textbf{Overview of the \methodname framework.} The benchmark couples a reproducible environment module, a GUI agent, an online user simulator grounded in user profiles and logs, and a hybrid evaluation pipeline combining rule based checks with LLM-as-a-judge scoring.}
\label{fig:method}
\end{figure*}

For personalized and proactive tasks, \methodname instantiates a user simulator $\pi_u$ to provide realistic interactive feedback (Figure~\ref{fig:method}). Each user is associated with two complementary components: a structured profile $P$, which encodes basic information together with personalized attributes such as preferences, habits, and constraints, and a timestamped interaction log $H$, which records prior on-device operations in the form of (time, location, action) entries. Concrete instances of $P$ and $H$ are provided in Appendix~\ref{app:user_profile}.
Crucially, $P$ and $H$ are asymmetrically distributed across the two agents. The profile $P$ is exclusively accessible to $\pi_u$, serving as hidden context that grounds its role play behavior, whereas the interaction log $H$ is exposed only to the GUI agent $\pi$, which must infer user preferences from observable behavioral patterns rather than from privileged profile knowledge. At each task, $\pi_u$ is conditioned on $P$, the current environment state $S$, and task specific instructions, enabling it to role play diverse users across varying profiles. When $\pi$ issues an \texttt{ask\_user} action, $\pi_u$ generates a response from a role grounded prompt constructed over $(P, S)$ and the dialogue history. This design supports evaluating whether agents can elicit user preferences in personalized tasks, and whether they exhibit appropriate initiative calibration and post-rejection restraint in proactive tasks.

% Figure~\ref{fig:method} provides an overview of the \methodname pipeline, including the environment module, GUI agent, online user simulator, and hybrid evaluation components.

\subsection{Task Definition}
\methodname comprises 42 general tasks, 86 personalized tasks, and 64 proactive tasks. Each task initializes the agent with a user instruction $g$. The input context additionally incorporates the exposed user logs $H$, and current environment state $S$ (e.g., current time and place) for personalized and proactive tasks. User profiles $P$ are defined across four roles---Researcher, Developer, Student, and Grandma---each characterized by name, age, work place and so on (see Figure~\ref{fig:method}). At each step $t$, the agent samples actions according to
\[
a_t \sim \pi(a\mid g,o_t,h_{<t},H,S,r_t), \qquad
a_t \in A.
\]
here $o_t$ is the current screenshot, $r_t$ is optional environment feedback (most notably the latest \texttt{ask\_user} response), and $h_{<t}$ is the past interaction history. Thus, unlike standard GUI agents that condition only on the instruction and screenshot, \methodname agents additionally receive history grounded textual context at initialization and may obtain user feedback during execution.
\paragraph{General Tasks}
General tasks are explicit instructions that require no inference over user-specific context. This subset serves as a baseline for assessing the agent's grounded GUI execution capability in isolation from preference reasoning and proactive decision-making.

\paragraph{Personalized Tasks}
Personalized tasks are ambiguous instructions whose against user-specific preferences encoded in $P$. For instance, an instruction such as ``\textit{order lunch for me today}'' implicitly requires the agent to determine the user's dietary preferences from $H$ or through interaction with $\pi_u$. When the agent issues a clarification question $m_t$ (i.e., $a_t=\texttt{ask\_user}$), the user simulator returns a free-form reply $r_t \sim \pi_u(\cdot \mid m_t, P, S)$ . Notably, templates are instantiated over task specific role subsets rather than a single globally fixed profile; the number of supported roles vary from one to four across templates.

\paragraph{Proactive Tasks}
Proactive tasks omit explicit instructions entirely: the agent receives only current state(time, location, and on-device GUI state) and must autonomously select one of three strategies---direct execution, proposing an action for confirmation, or remaining silent. For instance, after the user arrives at the office in the morning, the agent may order coffee, seek confirmation, or remain silent. If the agent seeks confirmation (i.e., $a_t=\texttt{ask\_user}$), the user simulator returns a response $r_t \sim \pi_u(\cdot \mid m_t, P, S)$ containing an explicit accept or reject decision regarding the proposed action. Each proactive template is evaluated across all four roles, so identical trigger conditions may yield different intervention decisions depending on the user's routine. The agent must infer whether to act, ask, or remain silent---and if it asks, condition its subsequent execution on $r_t$, proceeding upon acceptance or adjusting upon rejection.

% \subsection{User Logs Synthesis}
% Automated + Human Filtering
% \paragraph{APP Extending}
% \paragraph{Noise Synthesis}
\subsection{Hybrid Evaluation Strategy}
We adopt a hybrid evaluation strategy combining Rule-based and LLM-based Judges.
\paragraph{Rule-Based Judge}
The rule based component applies deterministic checks over verifiable states, including recipient correctness, event or order creation, alarm or setting configuration, time window validity, and trajectory level violations such as unsafe actions after user rejection. For fully programmatic tasks, it returns a binary signal $S_{\mathrm{rule}} \in \{0,1\}$. In a subset of hybrid personalized tasks, the same deterministic checks instead provide a bounded base score, which is later fused with the LLM judge.
\paragraph{LLM-as-a-judge}
The semantic component employs a rubric-conditioned judge that evaluates the extracted evidence and dialogue trace against a task-specific weighted rubric spanning dimensions such as preference alignment, trade-off quality, communication style, contextual appropriateness, and clarification quality. The judge returns both a normalized semantic score and a natural-language rationale, which we retain as the evaluation reason. The final score is
\[
S_i=\lambda_i S_{\mathrm{rule}}+(1-\lambda_i)S_{\mathrm{llm}},
\qquad
\lambda_i\in[0,1].
\]
We set $\lambda_i=1$ for fully deterministic tasks, $\lambda_i=0$ for purely semantic tasks. For personalized tasks, $\lambda_i$ is set in proportion to the share of preference dependent requirements in task $i$, such that tasks involving more personalized criteria assign greater weight to the LLM judge. The evaluator returns the final score along with a reason inherited from the active evaluation path---either the deterministic checker or the LLM judge.

\section{Experiment}
\label{sec:exp}
% 中文提纲（与当前代码协议对齐）：
% 1) Setup: 192 episodes = 42 general + 86 personalized + 64 proactive
% 2) Main results: matched General / Personalized / Proactive split-wise SR,
%    and Personalized additionally reports Avg. Score
% 3) Analysis: persona breakdown, clarification behavior, routine safety
% 4) Ablation: all vs rag, tfidf vs embedding, clean vs noise, top-k

\subsection{Experimental Setup}

\paragraph{Implementation Details.}
We evaluate two memory implementations: full history (\texttt{all}) and retrieved log snippets (\texttt{rag}), where the latter employs an embedding-based retriever with a variable retrieval budget~$k$. For both implementations, we further consider two log conditions: \emph{clean} logs, which retain only entries pertaining to user preferences, and \emph{noisy} logs, which additionally include irrelevant entries. Unless otherwise specified, all experiments adopt the \texttt{all} + \texttt{noisy} setting.
For interaction-needed tasks, we use \texttt{gpt-4o} as user simulator $\pi_u$ to produce role-grounded replies and accept/reject decisions.

\paragraph{Baselines and Metrics.}
We evaluate 11 state-of-the-art models in three categories: \textit{(1) GUI-specific models}, including MAI-UI-8B~\citep{zhou2025maiui}, UI-Venus-1.5-8B~\citep{gao2026uivenus1.5}, and GUI-Owl-1.5-8B~\citep{xu2026mobileagentv3.5}; \textit{(2) General open-source models}, including Qwen3-VL-8B~\citep{bai2025qwen3VL}, Qwen3-VL-32B~\citep{bai2025qwen3VL}, Qwen3.5-9B, Qwen3.5-122B-A10B, and Qwen3.5-397B-A17B. \textit{(3)
Closed-source models}, including Gemini 3.1 Pro Preview~\citep{team2023gemini},
Claude Sonnet 4.6, and Seed 2.0 Pro.

For task $i$, let $S_i \in [0,1]$ denote the task score, $s_i=\mathbb{I}[S_i>0.99]$ the binary success indicator, $t_i$ the number of executed actions, and $c_i$ the number of \texttt{ask\_user} queries. We organize our evaluation metrics into three tiers according to their scope of applicability.
\begin{itemize}
    \item Across \textbf{all} evaluation splits, we report \textbf{Success Rate (SR)}, defined as the proportion of tasks successfully completed within a split, and \textbf{Efficiency}, defined as $50 / \mathrm{AveSteps}(\mathcal{I})$, so that larger values consistently indicate more economical execution.
    \item For \textbf{personalized} tasks, we additionally report \textbf{Average Score}, defined as the mean instance-level score over all personalized examples. Unlike binary success, this metric captures partial preference alignment. Following the UIQ metric in MobileWorld~\citep{kong2025mobileworld}, we define \textbf{Interaction Efficiency (IE)} as
\[
\mathrm{IE}(\mathcal{I})=\frac{1}{|\mathcal{I}|}\sum_{i\in\mathcal{I}}\frac{S_i}{\max(c_i,1)},
\]
which measures the effectiveness of the agent interactions with users.
\item For \textbf{proactive} tasks, we report three policy-aware indicators computed over complementary subsets of instances. The \emph{Act} rate measures whether the agent intervenes when intervention is warranted, the \emph{Silent} rate measures whether the agent appropriately refrains from acting when intervention is unnecessary, and the \emph{Stop} rate measures whether the agent ceases further attempts after an explicit user rejection. Taken together, these metrics provide a comprehensive view of execution quality, action efficiency, preference alignment, clarification efficiency, and proactive restraint.
\end{itemize}

\definecolor{bestbg}{RGB}{255,242,204}
\definecolor{secondbg}{RGB}{221,235,247}
\newcommand{\best}[1]{{\textbf{#1}}}
\newcommand{\second}[1]{{\underline{#1}}}

\begin{table*}[t]
\centering
\small
\setlength{\tabcolsep}{3.6pt}
\renewcommand{\arraystretch}{1.10}
\caption{Main results on \methodname under the noisy full-history memory setting (\texttt{Full Log, Noisy}), where each agent receives the complete user logs together with irrelevant history. Each task type is split into easy and hard subsets, and \textbf{Overall SR} is computed over all tasks. General and Proactive columns report Success Rate (SR), while Personalized additionally reports Average Score. \textbf{Best} and \underline{second-best} denote the top two values in each column.}

\label{tab:main_results}
\begin{tabular}{lccccccccc}
\toprule
\multirow{3}{*}{Model} & \multirow{3}{*}{\makecell{Overall\\SR}} & \multicolumn{2}{c}{General} & \multicolumn{4}{c}{Personalized} & \multicolumn{2}{c}{Proactive} \\
\cmidrule(lr){3-4} \cmidrule(lr){5-8} \cmidrule(lr){9-10}
 & & easy & hard & \multicolumn{2}{c}{easy} & \multicolumn{2}{c}{hard} & easy & hard \\
\cmidrule(lr){3-3} \cmidrule(lr){4-4} \cmidrule(lr){5-6} \cmidrule(lr){7-8} \cmidrule(lr){9-9} \cmidrule(lr){10-10}
 & & SR & SR & SR & Score& SR & Score & SR & SR \\
\midrule
\rowcolor{gray!10}\multicolumn{10}{l}{\small\itshape Open-source models} \\
UI-Venus-1.5-8B   & 26.0 & 72.2 & 25.0 & 18.6 & 0.48 & 7.0  & 0.40 & 34.4 & 31.3 \\
Qwen3-VL-8B       & 21.9 & 72.2 & 4.2  & 7.0 & 0.27 & 7.0  & 0.25 & 46.9 & 21.9 \\
GUI-Owl-1.5-8B    & 22.4 & 77.8 & 33.3 & 9.3 & 0.42 & 2.4  & 0.34 & 28.1 & 21.9 \\
MAI-UI-8B         & 26.0 & \best{100.0} & 29.2 & 16.3 & 0.40 & 11.9 & 0.31 & 17.9 & 22.2 \\
Qwen3.5-122B-A10B & 27.1 & \second{94.4} & 25.0 & 30.2 & \second{0.69} & 9.5 & 0.60 & 25.0 & 12.5 \\
Qwen3-VL-32B      & 29.2 & 77.8 & 25.0 & 18.6 & 0.44 & 2.4  & 0.26 & 50.0 & 34.4 \\
Qwen3.5-9B        & 33.3 & 83.3 & 12.5 & 9.3  & 0.17 & 0.0  & 0.18 & 65.6 & \best{65.6} \\
Qwen3.5-397B-A17B & 37.5 & 83.3 & 20.8 & 25.6 & 0.59 & 2.3 & 0.48 & \second{68.8} & 56.3 \\
\addlinespace[2pt]
\midrule
\rowcolor{gray!10}\multicolumn{10}{l}{\small\itshape Closed-source models} \\
Gemini 3.1 Pro Preview & 44.3 & \second{94.4} & \second{66.7} & \second{34.9} & \best{0.78} & 20.9 & \second{0.75} & 50.0 & 38.9 \\
Seed 2.0 Pro      & \second{51.6} & \best{100.0} & 62.5 & 32.6 & 0.65 & \second{27.9} & 0.57 & 62.5 & \second{62.5} \\
Claude Sonnet 4.6 & \best{60.4} & \second{94.4} & \best{70.8} & \best{44.2} & \best{0.78} & \best{44.2} & \best{0.80} & \best{84.4} & 53.1 \\
\bottomrule
\end{tabular}
\end{table*}

\subsection{Main Results}
\paragraph{Difficulty Progression Across Task Types.}
Table~\ref{tab:main_results} reveals a clear progression in difficulty, from explicit GUI execution to personalized assistance and finally proactive service. In the easy general split, MAI-UI-8B and Seed 2.0 Pro both achieve a success rate of 100.0\%. This suggests that executing fully specified instructions is no longer the primary bottleneck. However, performance declines sharply once tasks require user-specific reasoning. On the hard personalized split, Claude Sonnet 4.6 attains a success rate of 44.2\%, whereas all open-source models remain below 12\%. At the same time, the average score is consistently much higher than strict success rate on personalized tasks, suggesting that many agents can partially infer user preferences, yet still fail to translate that partial alignment into fully correct end-to-end behavior. Proactive tasks show a different pattern: model rankings are less stable across difficulty levels, and models such as Qwen3.5-9B remain competitive despite weak personalized performance. This indicates that proactive calibration is not simply another form of preference disambiguation. Overall, closed-source models still lead the table, with Claude Sonnet 4.6 achieving the best overall success rate of 60.4\%. However, the substantial gap between general execution and the personalized and proactive settings shows that profile grounding and calibrated initiative remain unsolved.

\begin{figure*}[t]
\centering
\includegraphics[width=\textwidth]{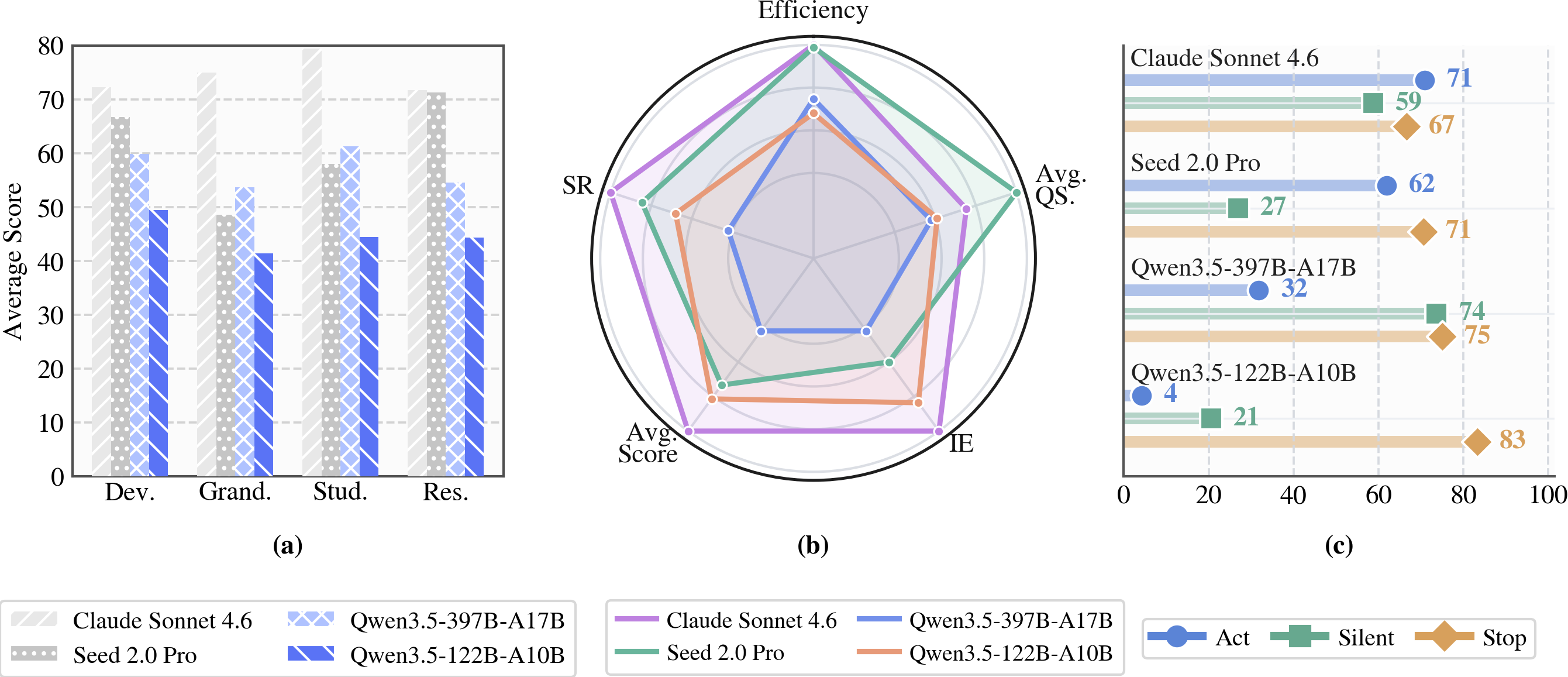}
\caption[Visualization analyses]{Visualization analyses. (a) Average score across four user roles: Developer (Dev.), Grandma (Grand.), Student (Stud.), and Researcher (Res.). (b) Personalized interaction metrics, including Efficiency (defined as $50/\text{Avg.\ Steps}$), Average Queries, and Interaction Efficiency (IE). (c) Proactive safety rates, including Act, Silent, and Stop.}
\label{fig:analysis_panels}
\end{figure*}

\paragraph{Role Dependence.}
Figure~\ref{fig:analysis_panels}(a) shows that performance remains sensitive to user role. Claude Sonnet 4.6 leads on all four roles and stays relatively stable at 71.7\%--79.4\%, while Seed 2.0 Pro varies much more, rising to 71.3\% on the researcher role but dropping to 48.5\% on the grandma role. Across models, grandma is the hardest role on average, and student produces the largest spread. This supports our core motivation: the challenge is not generic task completion, but whether the agent can make decisions that fit the personalized needs of different users.

\paragraph{Preference Acquisition Through Interaction.}
Figure~\ref{fig:analysis_panels}(b) shows that better personalization is not simply a matter of asking more questions. Claude Sonnet 4.6 achieves the strongest overall profile, with a 44.2\% success rate and a 78.9\% average score while asking only 0.4 questions per task on average. By contrast, Seed 2.0 Pro asks about twice as many questions, yet still lags behind, which suggests that interaction helps only when the acquired preference signal is turned into better downstream actions. The two Qwen models reinforce the same point: they ask almost the same number of questions, but Qwen3.5-122B-A10B achieves noticeably better scores, while both still require more than 36 steps on average. The key bottleneck is therefore not whether the agent asks, but whether it can efficiently translate user feedback into correct end-to-end execution.

\paragraph{Proactive Safety Analysis: Initiative versus Restraint.}
Figure~\ref{fig:analysis_panels}(c) shows that proactive service is fundamentally a calibration problem. Claude Sonnet 4.6 is the most balanced model, with the best Act score at 70.8\% and competitive performance on the other two metrics. Qwen3.5-397B-A17B shows the opposite profile, leading on Silent at 73.7\% and reaching 75.0\% on Stop, but dropping to 31.8\% on Act. Qwen3.5-122B-A10B pushes this tradeoff even further, with the best Stop score at 83.3\% but very weak Act and Silent performance. The main insight is that proactive ability cannot be summarized by a single safety score: an effective agent must know when to intervene, when to stay silent, and when to back off after rejection.

\subsection{Ablation Studies}
\paragraph{Memory Implementation Matters.}
Beyond downstream action generation, \methodname also evaluates how agents access long term user evidence. Table~\ref{tab:memory_results} compares three agents under four memory configurations: full log and RAG log, each in clean and noisy variants. The central finding is that the optimal memory interface is model dependent rather than universal. \texttt{Qwen3-VL-8B} benefits substantially from selective retrieval, improving from 13.6\% (full log clean) to 20.4\% (RAG log clean), suggesting that compact evidence exposure sharpens preference grounding. In contrast, \texttt{UI-Venus-1.5-8B} performs better with full log access, indicating that aggressive compression can discard useful context for certain architectures. \texttt{MAI-UI-8B} remains weak across all settings and degrades further under RAG noisy (9.3\%), revealing that noisy retrieval can destabilize fragile memory utilization. These results underscore that robust personalization requires not only capable GUI execution but also careful design of how user logs are surfaced and filtered.

\begin{figure*}[t]
\centering
\includegraphics[width=0.90\textwidth]{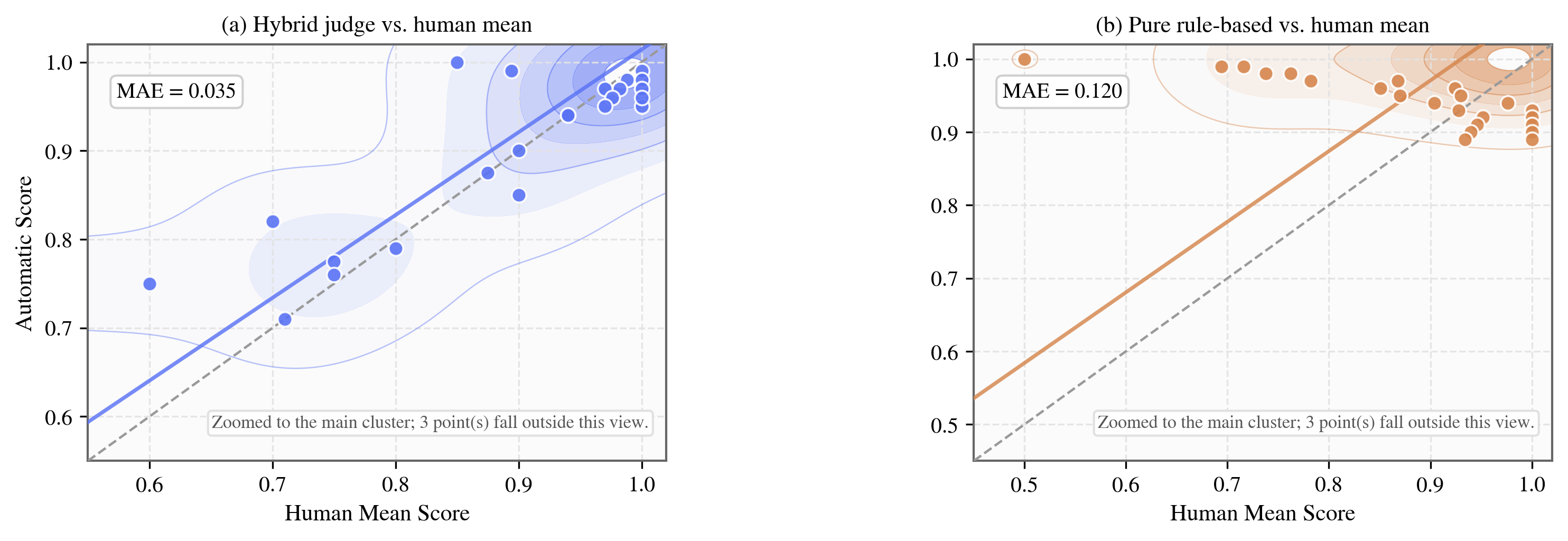}
\caption[Judge sensitivity against human ratings]{\textbf{Judge sensitivity against human ratings.} Task-level scatter plots comparing two automatic evaluators against the mean score of four human experts on 26 shared trajectories. Each point denotes one task, the dashed diagonal indicates perfect agreement, and the inset reports mean absolute error. The hybrid judge (LLM-as-a-judge combined with rule-based scoring) exhibits tighter clustering around the diagonal and lower error than the pure rule-based variant, confirming stronger alignment with human judgment.}
\label{fig:judge_sensitivity}
\end{figure*}

\paragraph{Judge and Simulator Sensitivity.}
To validate the evaluation protocol, we fix 26 task trajectories and compare automatic scores against mean ratings from four human experts. As shown in Figure~\ref{fig:judge_sensitivity}, the hybrid evaluator (LLM-as-a-judge combined with rule-based scoring) achieves a lower mean absolute error and tighter clustering around the perfect-agreement diagonal than the pure rule-based variant. This confirms the complementarity of both components: deterministic rules preserve verifiability on hard constraints, while the LLM judge captures semantic dimensions such as preference satisfaction that resist manual encoding, yielding a more human-aligned evaluation overall.

\subsection{Discussion}

\paragraph{Error Analysis.}
% !TEX root = ../main.tex
\begin{wraptable}{r}{0.48\columnwidth}
\vspace{-0.45\baselineskip}
\centering
\small
\setlength{\tabcolsep}{2.2pt}
\renewcommand{\arraystretch}{1.05}
\caption{Overall success rate under four memory settings, computed over personalized and proactive tasks only.}
\label{tab:memory_results}
\begin{tabular}{lcccc}
\toprule
\multirow{2}{*}{Model} & \multicolumn{2}{c}{Full Log} & \multicolumn{2}{c}{RAG Log} \\
\cmidrule(lr){2-3} \cmidrule(lr){4-5}
 & Clean & Noisy & Clean & Noisy \\
\midrule
MAI-UI-8B & 11.1 & 13.6 & 12.3 & 9.3 \\
Qwen3-VL-8B & 13.6 & 17.2 & 20.4 & 19.8 \\
UI-Venus-1.5-8B & 15.6 & 20.3 & 13.7 & 19.6 \\
\bottomrule
\end{tabular}
\vspace{-0.25\baselineskip}
\end{wraptable}

\begin{figure*}[t]
\centering
\includegraphics[width=\textwidth]{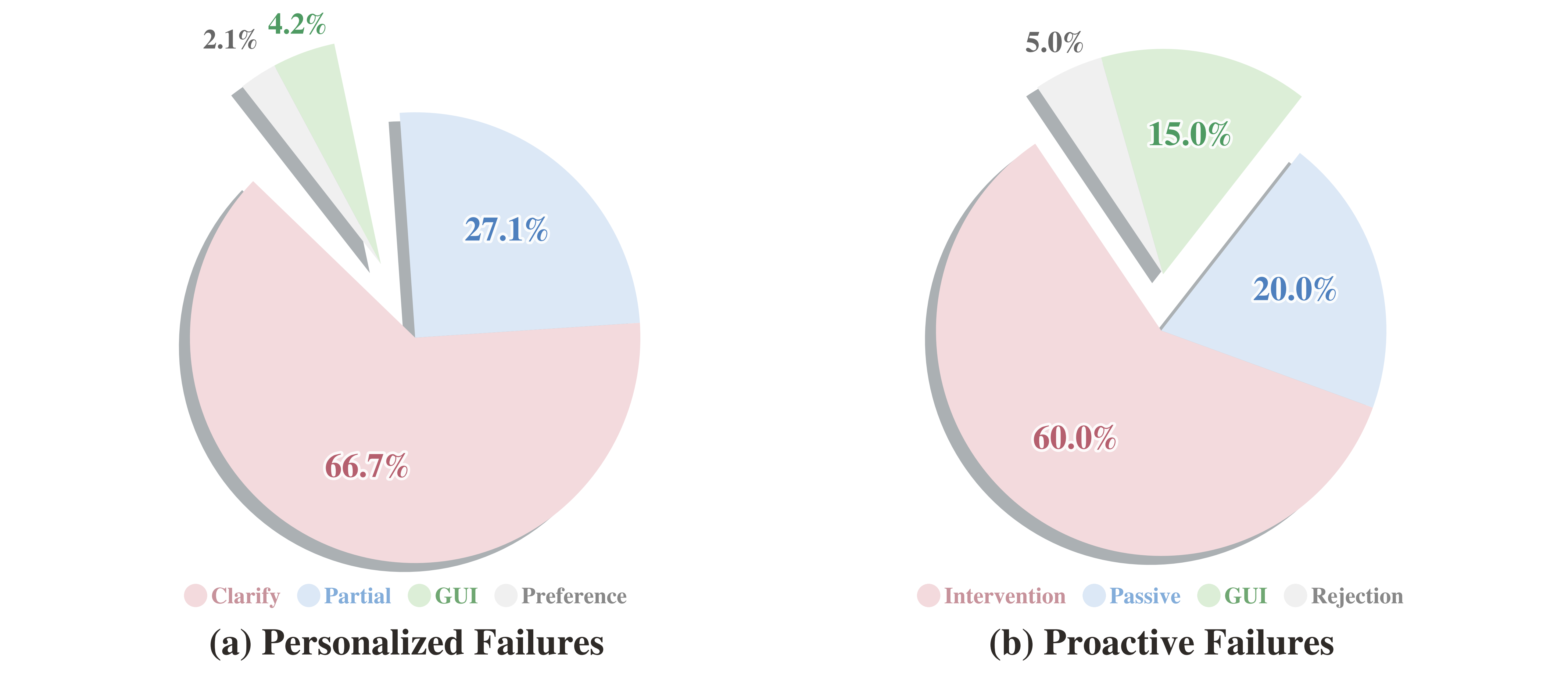}
\caption[Failure mode breakdown]{Failure mode breakdown. (a) Personalized failures are categorized into Clarify (insufficient clarification), Partial (partial preference satisfaction), Preference (preference misidentification), and GUI (GUI navigation failure). Most failures come from Clarify and Partial. (b) Proactive failures are categorized into Intervention (unwarranted intervention), Passive (false passivity), GUI (GUI navigation failure), and Rejection (post-rejection violation).}
\label{fig:error_analysis}
\end{figure*}

To understand why agents fail on personalized and proactive tasks, we manually categorize all failure trajectories produced by Claude Sonnet 4.6; the results are shown in Figure~\ref{fig:error_analysis}.

For personalized tasks (Figure~\ref{fig:error_analysis}(a)), failures are dominated by Clarify errors (66.7\%), with Partial failures (27.1\%) as the second largest category, while GUI (4.2\%) and Preference (2.1\%) errors are rare. A key insight is that current models still struggle to acquire user preferences effectively through interaction: the fact that insufficient clarification accounts for the majority of failures suggests that the model often does not ask the right follow-up questions before acting. The substantial share of Partial failures further shows that even when the main preference is identified, the model often fails to compose multiple constraints correctly.

For proactive tasks (Figure~\ref{fig:error_analysis}(b)), Intervention errors account for the majority of failures (60.0\%), followed by Passive (20.0\%), GUI (15.0\%), and Rejection (5.0\%). This suggests that proactive failure is primarily a calibration problem rather than an execution problem: Intervention and Passive together make up 80.0\% of all failures, far exceeding downstream GUI errors. Moreover, the much higher rate of Intervention than Passive suggests that current agents are more prone to over-act than to miss opportunities for action.

Overall, the two settings expose different bottlenecks. Personalized tasks are limited mainly by interactive preference acquisition and multi-constraint preference composition, whereas proactive tasks are limited mainly by initiative calibration. This points to different priorities for future agents: stronger interactive preference elicitation and compositional preference modeling for personalization, and better trigger calibration, abstention, and rejection-aware decision policies for proactivity.

\section{Conclusion}
\methodname targets a missing part of mobile agent evaluation: the ability to \emph{act as the right assistant for the right user}, rather than merely execute explicit instructions. By combining a reproducible Android emulator environment, structured profiles, user logs, user interaction, and hybrid evaluation, \methodname turns personalization from an offline intent-alignment problem into an online execution-grounded benchmark.

Our experiments show that current agents still fall far short of this goal. Even the strongest models exhibit a large gap between explicit-task execution and personalized decision making, and the gap becomes even larger in proactive routine scenarios that require initiative calibration and restraint after rejection. In other words, existing models can often navigate the interface, but they still struggle to decide \emph{what} should be done for \emph{which} user and \emph{when} it should be done.

We hope \methodname can serve both as a benchmark and as a research platform for future work on personalized mobile intelligence. Beyond improving execution accuracy, we believe the next major advances will come from better long-term memory access, stronger ambiguity-resolution policies, and safer proactive decision boundaries. These are the ingredients required for turning mobile agents from competent GUI operators into trustworthy personal assistants.

\clearpage

\bibliography{main,custom}

@misc{rawles2024androidworld,
  title = {{AndroidWorld}: A Dynamic Benchmarking Environment for Autonomous Agents},
  author = {Rawles, Christopher and Clinckemaillie, Sarah and Chang, Yifan and Waltz, Jonathan and Lau, Gabrielle and Fair, Marybeth and Li, Alice and Bishop, William and Li, Wei and Campbell-Ajala, Folawiyo and Toyama, Daniel and Berry, Robert and Tyamagundlu, Divya and Lillicrap, Timothy and Riva, Oriana},
  year = 2024,
  month = may,
  number = {arXiv:2405.14573},
  eprint = {2405.14573},
  primaryclass = {cs},
  publisher = {arXiv},
  doi = {10.48550/arXiv.2405.14573},
  urldate = {2026-03-12},
  archiveprefix = {arXiv}
}

@inproceedings{xu2025androidlab,
  title = {{AndroidLab}: Training and Systematic Benchmarking of Android Autonomous Agents},
  shorttitle = {AndroidLab},
  booktitle = {Proceedings of the 63rd Annual Meeting of the Association for Computational Linguistics (Volume 1: Long Papers)},
  author = {Xu, Yifan and Liu, Xiao and Sun, Xueqiao and Cheng, Siyi and Yu, Hao and Lai, Hanyu and Zhang, Shudan and Zhang, Dan and Tang, Jie and Dong, Yuxiao},
  year = 2025,
  pages = {2144--2166},
  publisher = {Association for Computational Linguistics},
  address = {Vienna, Austria},
  doi = {10.18653/v1/2025.acl-long.107},
  urldate = {2026-03-12},
  langid = {english}
}

@inproceedings{chen2024spabench,
  title = {{SPA-Bench}: A Comprehensive Benchmark for SmartPhone Agent Evaluation},
  shorttitle = {SPA-Bench},
  booktitle = {The Thirteenth International Conference on Learning Representations},
  author = {Chen, Jingxuan and Yuen, Derek and Xie, Bin and Yang, Yuhao and Chen, Gongwei and Wu, Zhihao and Li, Yixing and Zhou, Xurui and Liu, Weiwen and Wang, Shuai and Zhou, Kaiwen and Shao, Rui and Nie, Liqiang and Wang, Yasheng and Hao, Jianye and Wang, Jun and Shao, Kun},
  year = 2025,
  note = {Spotlight},
  urldate = {2026-03-12},
  langid = {english}
}

@misc{yan2025androiddaily,
  title = {Step-GUI Technical Report},
  shorttitle = {AndroidDaily},
  author = {Yan, Haolong and Wang, Jia and Huang, Xin and Shen, Yeqing and Meng, Ziyang and Fan, Zhimin and Tan, Kaijun and Gao, Jin and Shi, Lieyu and others},
  year = 2025,
  month = dec,
  number = {arXiv:2512.15431},
  eprint = {2512.15431},
  primaryclass = {cs},
  publisher = {arXiv},
  doi = {10.48550/arXiv.2512.15431},
  urldate = {2026-03-12},
  archiveprefix = {arXiv},
  note = {Introduces the AndroidDaily benchmark}
}

@misc{kong2025mobileworld,
  title = {{MobileWorld}: Benchmarking Autonomous Mobile Agents in Agent-User Interactive and MCP-Augmented Environments},
  shorttitle = {MobileWorld},
  author = {Kong, Quyu and Zhang, Xu and Yang, Zhenyu and Gao, Nolan and Liu, Chen and Tong, Panrong and Cai, Chenglin and Zhou, Hanzhang and Zhang, Jianan and Chen, Liangyu and Liu, Zhidan and Hoi, Steven and Wang, Yue},
  year = 2025,
  month = dec,
  number = {arXiv:2512.19432},
  eprint = {2512.19432},
  primaryclass = {cs},
  publisher = {arXiv},
  doi = {10.48550/arXiv.2512.19432},
  urldate = {2026-03-12},
  archiveprefix = {arXiv}
}

@misc{yang2025fingertip20k,
  title = {FingerTip 20K: A Benchmark for Proactive and Personalized Mobile LLM Agents},
  shorttitle = {FingerTip 20K},
  author = {Yang, Qinglong and Li, Haoming and Zhao, Haotian and Yan, Xiaokai and Ding, Jingtao and Xu, Fengli and Li, Yong},
  year = 2025,
  month = jun,
  number = {arXiv:2507.21071},
  eprint = {2507.21071},
  primaryclass = {cs.HC},
  publisher = {arXiv},
  doi = {10.48550/arXiv.2507.21071},
  urldate = {2026-03-12},
  archiveprefix = {arXiv}
}

@misc{lyu2026personalalign,
  title = {PersonalAlign: Hierarchical Implicit Intent Alignment for Personalized GUI Agent with Long-Term User-Centric Records},
  shorttitle = {PersonalAlign},
  author = {Lyu, Yibo and Chen, Gongwei and Shao, Rui and Guan, Weili and Nie, Liqiang},
  year = 2026,
  month = jan,
  number = {arXiv:2601.09636},
  eprint = {2601.09636},
  primaryclass = {cs.AI},
  publisher = {arXiv},
  doi = {10.48550/arXiv.2601.09636},
  urldate = {2026-03-12},
  archiveprefix = {arXiv}
}

@misc{kong2026proactivemobile,
  title = {ProactiveMobile: A Comprehensive Benchmark for Boosting Proactive Intelligence on Mobile Devices},
  shorttitle = {ProactiveMobile},
  author = {Kong, Dezhi and Feng, Zhengzhao and Liang, Qiliang and Wang, Hao and Sun, Haofei and Yang, Changpeng and Li, Yang and Zhou, Peng and Nie, Shuai and Wang, Hongzhen and Zhou, Linfeng and Jia, Hao and Xu, Jiaming and Shi, Runyu and Huang, Ying},
  year = 2026,
  month = feb,
  number = {arXiv:2602.21858},
  eprint = {2602.21858},
  primaryclass = {cs.AI},
  publisher = {arXiv},
  doi = {10.48550/arXiv.2602.21858},
  urldate = {2026-03-12},
  archiveprefix = {arXiv}
}

@misc{chai2026pirabench,
  title = {PIRA-Bench: A Transition from Reactive GUI Agents to GUI-based Proactive Intent Recommendation Agents},
  shorttitle = {PIRA-Bench},
  author = {Chai, Yuxiang and Tang, Shunye and Xiao, Han and Liu, Rui and Li, Hongsheng},
  year = 2026,
  month = mar,
  number = {arXiv:2603.08013},
  eprint = {2603.08013},
  primaryclass = {cs.AI},
  publisher = {arXiv},
  doi = {10.48550/arXiv.2603.08013},
  urldate = {2026-03-12},
  archiveprefix = {arXiv}
}

@misc{wang2026meagent,
  title = {Me-Agent: A Personalized Mobile Agent with Two-Level User Habit Learning for Enhanced Interaction},
  shorttitle = {Me-Agent},
  author = {Wang, Shuoxin and Liu, Chang and Loo, Gowen and Zheng, Lifan and Wei, Kaiwen and Zeng, Xinyi and Zhang, Jingyuan and Tian, Yu},
  year = 2026,
  month = jan,
  number = {arXiv:2601.20162},
  eprint = {2601.20162},
  primaryclass = {cs.CL},
  publisher = {arXiv},
  doi = {10.48550/arXiv.2601.20162},
  urldate = {2026-03-12},
  archiveprefix = {arXiv}
}

@misc{rawles2023aitw,
  title   = {Android in the Wild: A Large-Scale Dataset for Android Device Control},
  author  = {Rawles, Christopher and Li, Alice and Rodriguez, Daniel and Riva, Oriana and Lillicrap, Timothy},
  year    = 2023,
  eprint  = {2307.10088},
  archiveprefix = {arXiv},
  primaryclass  = {cs.CV}
}

@misc{li2024androidcontrol,
  title   = {On the Effects of Data Scale on UI Control Agents},
  author  = {Li, Wei and Bishop, William and Li, Alice and Rawles, Christopher and Riva, Oriana and Lillicrap, Timothy},
  year    = 2024,
  eprint  = {2406.03679},
  archiveprefix = {arXiv},
  primaryclass  = {cs.AI}
}

@article{liu2026memgui,
  title={MemGUI-Bench: Benchmarking Memory of Mobile GUI Agents in Dynamic Environments},
  author={Liu, Guangyi and Zhao, Pengxiang and Liang, Yaozhen and Luo, Qinyi and Tang, Shunye and Chai, Yuxiang and Lin, Weifeng and Xiao, Han and Wang, WenHao and Chen, Siheng and others},
  journal={arXiv preprint arXiv:2602.06075},
  year={2026}
}

@inproceedings{lu2026uir1,
  title={Ui-r1: Enhancing efficient action prediction of gui agents by reinforcement learning},
  author={Lu, Zhengxi and Chai, Yuxiang and Guo, Yaxuan and Yin, Xi and Liu, Liang and Wang, Hao and Xiao, Han and Ren, Shuai and Zhao, Pengxiang and Liu, Guangyi and others},
  booktitle={Proceedings of the AAAI Conference on Artificial Intelligence},
  volume={40},
  number={21},
  pages={17608--17616},
  year={2026}
}

@article{tang2025guig2,
  title={GUI-G$^2$: Gaussian Reward Modeling for GUI Grounding},
  author={Tang, Fei and Gu, Zhangxuan and Lu, Zhengxi and Liu, Xuyang and Shen, Shuheng and Meng, Changhua and Wang, Wen and Zhang, Wenqi and Shen, Yongliang and Lu, Weiming and others},
  journal={arXiv preprint arXiv:2507.15846},
  year={2025}
}

@article{ye2025mobileagentv3,
  title={Mobile-agent-v3: Fundamental agents for gui automation},
  author={Ye, Jiabo and Zhang, Xi and Xu, Haiyang and Liu, Haowei and Wang, Junyang and Zhu, Zhaoqing and Zheng, Ziwei and Gao, Feiyu and Cao, Junjie and Lu, Zhengxi and others},
  journal={arXiv preprint arXiv:2508.15144},
  year={2025}
}

@article{liu2025guisurvey,
  title={Llm-powered gui agents in phone automation: Surveying progress and prospects},
  author={Liu, Guangyi and Zhao, Pengxiang and Liang, Yaozhen and Liu, Liang and Guo, Yaxuan and Xiao, Han and Lin, Weifeng and Chai, Yuxiang and Han, Yue and Ren, Shuai and others},
  journal={arXiv preprint arXiv:2504.19838},
  year={2025}
}

@article{tang2025guisurvey,
  title={A survey on (m) llm-based gui agents},
  author={Tang, Fei and Xu, Haolei and Zhang, Hang and Chen, Siqi and Wu, Xingyu and Shen, Yongliang and Zhang, Wenqi and Hou, Guiyang and Tan, Zeqi and Yan, Yuchen and others},
  journal={arXiv preprint arXiv:2504.13865},
  year={2025}
}

@inproceedings{wu2026gem,
  title={Gem: Gaussian embedding modeling for out-of-distribution detection in gui agents},
  author={Wu, Zheng and Cheng, Pengzhou and Wu, Zongru and Dong, Lingzhong and Zhang, Zhuosheng},
  booktitle={Proceedings of the AAAI Conference on Artificial Intelligence},
  volume={40},
  number={40},
  pages={33989--33997},
  year={2026}
}

@article{qin2025uitars,
  title={Ui-tars: Pioneering automated gui interaction with native agents},
  author={Qin, Yujia and Ye, Yining and Fang, Junjie and Wang, Haoming and Liang, Shihao and Tian, Shizuo and Zhang, Junda and Li, Jiahao and Li, Yunxin and Huang, Shijue and others},
  journal={arXiv preprint arXiv:2501.12326},
  year={2025}
}

@article{zhou2025maiui,
  title={MAI-UI Technical Report: Real-World Centric Foundation GUI Agents},
  author={Zhou, Hanzhang and Zhang, Xu and Tong, Panrong and Zhang, Jianan and Chen, Liangyu and Kong, Quyu and Cai, Chenglin and Liu, Chen and Wang, Yue and Zhou, Jingren and others},
  journal={arXiv preprint arXiv:2512.22047},
  year={2025}
}

@article{gu2025uivenus,
  title={Ui-venus technical report: Building high-performance ui agents with rft},
  author={Gu, Zhangxuan and Zeng, Zhengwen and Xu, Zhenyu and Zhou, Xingran and Shen, Shuheng and Liu, Yunfei and Zhou, Beitong and Meng, Changhua and Xia, Tianyu and Chen, Weizhi and others},
  journal={arXiv preprint arXiv:2508.10833},
  year={2025}
}

@article{gao2026uivenus1.5,
  title={UI-Venus-1.5 Technical Report},
  author={Gao, Changlong and Gu, Zhangxuan and Liu, Yulin and Qiu, Xinyu and Shen, Shuheng and Wen, Yue and Xia, Tianyu and Xu, Zhenyu and Zeng, Zhengwen and Zhou, Beitong and others},
  journal={arXiv preprint arXiv:2602.09082},
  year={2026}
}

@article{bai2025qwen3VL,
  title={Qwen3-vl technical report},
  author={Bai, Shuai and Cai, Yuxuan and Chen, Ruizhe and Chen, Keqin and Chen, Xionghui and Cheng, Zesen and Deng, Lianghao and Ding, Wei and Gao, Chang and Ge, Chunjiang and others},
  journal={arXiv preprint arXiv:2511.21631},
  year={2025}
}

@article{xu2026mobileagentv3.5,
  title={Mobile-Agent-v3. 5: Multi-platform Fundamental GUI Agents},
  author={Xu, Haiyang and Zhang, Xi and Liu, Haowei and Wang, Junyang and Zhu, Zhaozai and Zhou, Shengjie and Hu, Xuhao and Gao, Feiyu and Cao, Junjie and Wang, Zihua and others},
  journal={arXiv preprint arXiv:2602.16855},
  year={2026}
}

@article{team2023gemini,
  title={Gemini: a family of highly capable multimodal models},
  author={Team, Gemini and Anil, Rohan and Borgeaud, Sebastian and Alayrac, Jean-Baptiste and Yu, Jiahui and Soricut, Radu and Schalkwyk, Johan and Dai, Andrew M and Hauth, Anja and Millican, Katie and others},
  journal={arXiv preprint arXiv:2312.11805},
  year={2023}
}

@article{lu2025uis1,
  title={Ui-s1: Advancing gui automation via semi-online reinforcement learning},
  author={Lu, Zhengxi and Ye, Jiabo and Tang, Fei and Shen, Yongliang and Xu, Haiyang and Zheng, Ziwei and Lu, Weiming and Yan, Ming and Huang, Fei and Xiao, Jun and others},
  journal={arXiv preprint arXiv:2509.11543},
  year={2025}
}

@article{nathani2026proactive,
  title={Proactive Agent Research Environment: Simulating Active Users to Evaluate Proactive Assistants},
  author={Nathani, Deepak and Zhang, Cheng and Huan, Chang and Shan, Jiaming and Yang, Yinfei and Patel, Alkesh and Gan, Zhe and Wang, William Yang and Saxon, Michael and Wang, Xin Eric},
  journal={arXiv preprint arXiv:2604.00842},
  year={2026}
}

@misc{openclaw2026,
  title        = {OpenClaw},
  author       = {{OpenClaw}},
  year         = {2026},
  howpublished = {\url{https://github.com/openclaw/openclaw}},
  note         = {Open-source personal AI assistant, version 2026.3.8, accessed 2026-03-09}
}
\bibliographystyle{main}

\clearpage

\appendix
\tableofcontents

\clearpage

\section{Framework Pipeline}
\label{app:pipeline}

\begin{figure*}[h]
\centering
\includegraphics[width=\textwidth]{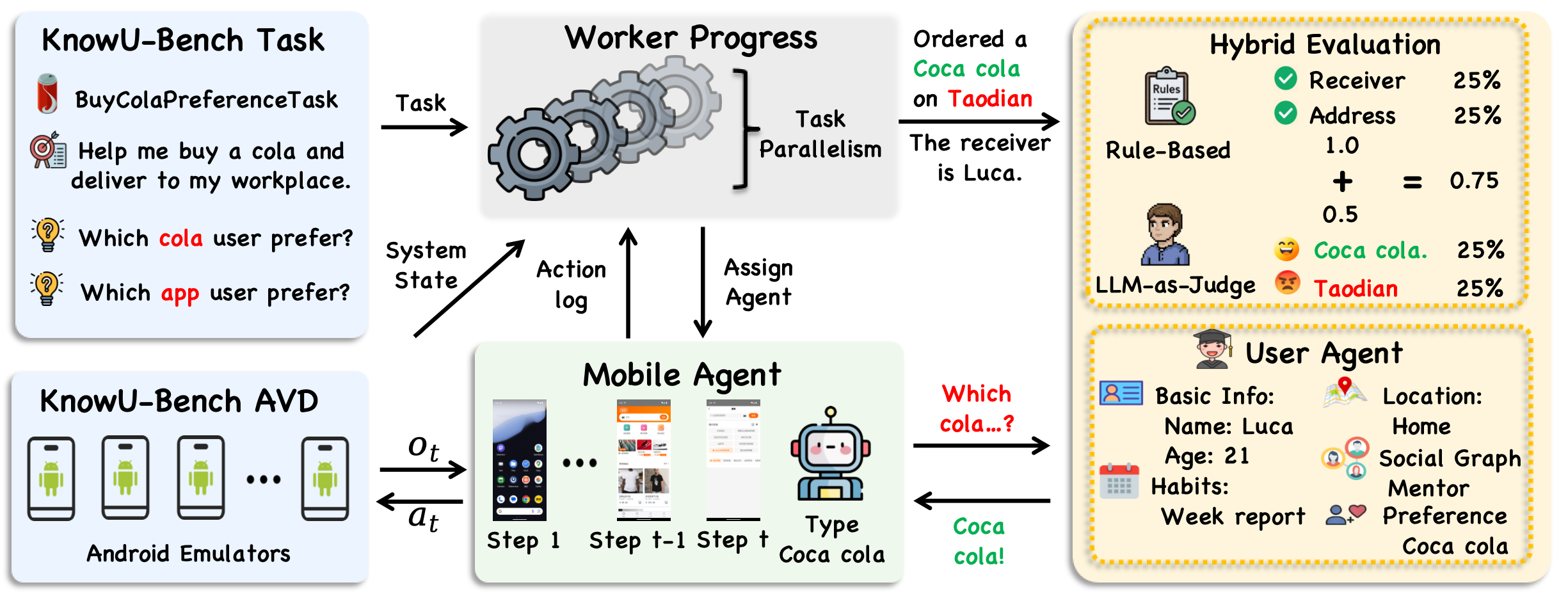}
\caption[Additional framework pipeline]{Additional view of the \methodname pipeline, showing task initialization, agent interaction, user simulation, and hybrid evaluation.}
\label{fig:appendix_pipeline}
\end{figure*}

Figure~\ref{fig:appendix_pipeline} provides an additional view of the end-to-end benchmark pipeline.

\section{GUI Action Space}
\label{app:action_space}

Table~\ref{tab:action_space} summarizes the GUI action space used by \methodname.

\begin{table}[h]
\centering
\small
\begin{tabularx}{\linewidth}{|l|l|X|}
\hline
\textbf{Action} & \textbf{Parameters} & \textbf{Description} \\
\hline
\texttt{click} & \texttt{x, y} & Tap at the specified coordinates \\
\hline
\texttt{double\_tap} & \texttt{x, y} & Double-tap at the specified coordinates \\
\hline
\texttt{long\_press} & \texttt{x, y} & Long-press at the specified coordinates \\
\hline
\texttt{drag} & \texttt{start\_x, start\_y, end\_x, end\_y} & Drag from start to end coordinates \\
\hline
\texttt{input\_text} & \texttt{text} & Type text into the focused field \\
\hline
\texttt{scroll} & \texttt{direction} & Scroll in the specified direction (up/down/left/right) \\
\hline
\texttt{navigate\_home} & --- & Return to the home screen \\
\hline
\texttt{navigate\_back} & --- & Navigate to the previous screen \\
\hline
\texttt{keyboard\_enter} & --- & Press the enter key \\
\hline
\texttt{wait} & --- & Wait for screen content to update \\
\hline
\texttt{answer} & \texttt{text} & Provide a textual response to the user (for IR tasks) \\
\hline
\texttt{status} & \texttt{goal\_status} & Mark task as complete or infeasible \\
\hline
\texttt{ask\_user} & \texttt{text} & Request clarification from the user \\
\hline
\end{tabularx}
\caption{Action Space}
\label{tab:action_space}
\end{table}

\section{App Information}
\label{app:information}
\subsection{App List}
% Summarize all covered apps, their domains, and their roles in general,
% personalized, and proactive tasks.
Table~\ref{tab:app_list} summarizes the apps covered by \methodname, including their functional roles, comparable commercial apps, and associated task counts.

% !TEX root = ../main.tex
\begin{table}[H]
\centering
\small
\caption{App coverage of KnowU-Bench. \#Tasks counts app level participations rather than unique episodes; each cross app task is counted for every involved app.}
\label{tab:app_list}
\begin{tabularx}{\linewidth}{lX>{\centering\arraybackslash}p{0.27\linewidth}r}
\toprule
\textbf{App} & \textbf{Description} & \textbf{Comparable Commercial App} & \textbf{\#Tasks} \\
\midrule
jingdian   & E-commerce shopping platform & JD.com & 35 \\
Taodian    & E-commerce shopping platform & Taobao & 35 \\
Messages   & SMS and chat messaging & - & 26 \\
Mattermost & Team collaboration and messaging & Slack & 25 \\
Settings   & System configuration & - & 20 \\
Calendar   & Manage events and schedules & Google Calendar & 18 \\
Maps       & Navigation and location services & Google Maps & 17 \\
Mastodon   & Decentralized social network & Twitter/X & 17 \\
chilemei   & Food ordering and delivery & Ele.me & 15 \\
Chrome     & Web browser for internet browsing & - & 15 \\
Contacts   & Manage contact information & - & 15 \\
Files      & File manager for device storage & - & 15 \\
Mail       & Email client for messaging & Gmail & 15 \\
tuantuan   & Food ordering and delivery & Meituan & 15 \\
Gallery    & View and manage photos & - & 13 \\
Clock      & Alarms, timers, and world clock & - & 7 \\
Docreader  & View and read documents & Adobe Reader & 5 \\
\bottomrule
\end{tabularx}
\end{table}

\subsection{App Coverage Expansion}
\label{app:coverage_expansion}
% Describe how \methodname extends the original MobileWorld app ecosystem
% and why the added apps are necessary for personalized evaluation.
Following the environment construction philosophy of MobileWorld~\cite{kong2025mobileworld}, we expand the original app ecosystem with four service oriented applications: two shopping apps (\texttt{Taodian} and \texttt{jingdian}) and two food delivery apps (\texttt{chilemei} and \texttt{tuantuan}). These applications provide controlled environments for preference sensitive service tasks, including platform choice, payment habit, delivery address selection, cuisine preference, and app specific ordering routines.

\textbf{Shopping apps.} Our shopping environments are adapted from the \texttt{mall\_fork} codebase\footnote{GitHub repository: \texttt{qykong/mall\_fork}.}, which itself derives from the Mall4Uni ecosystem. We retain the core shopping workflow while replacing backend dependencies with editable local mock data for products, user profiles, and delivery addresses. \texttt{jingdian} is constructed as a companion platform to \texttt{Taodian} with modified homepage layouts, product inventories, and visual styling, enabling evaluation of cross platform shopping preferences rather than behavior tied to a single interface.

\textbf{Food delivery apps.} Our delivery environments are built from the Flash Waimai project\footnote{GitHub repository released by Microapp Store.}. To make the environment self contained and reproducible, we remove the original backend dependent logic and convert the ordering workflow into a pure frontend pipeline backed by static shop, menu, rating, and address data. \texttt{chilemei} and \texttt{tuantuan} share the same basic interaction flow but differ in storefront content and UI appearance, allowing us to vary app surface realization while preserving controllable task semantics.

\textbf{Evaluation and deployment.} For all four service apps, we instrument critical completion events, especially successful order submission, with callback hooks that send structured order payloads to the host environment for automated verification. During deployment, we found the original UniApp based Android packaging unreliable in our emulator setup, particularly under x86\_64 related compatibility constraints. We therefore adopt a two stage pipeline that first compiles each app into a static H5 site and then packages it with Capacitor, together with cleartext HTTP support for host side callback APIs. This design preserves realistic interaction flows while making the expanded app suite substantially more stable and reproducible in the benchmark environment.

\section{User Profiles and Logs}
\label{app:user_profile}
\subsection{User Profiles}
\methodname stores each role profile as a YAML file. The current release includes four concrete profiles corresponding to the Developer, Grandma, Student, and Researcher roles. Although the concrete values differ substantially across roles, all profiles expose a unified top level interface so that tasks, simulators, and evaluators can access role information through the same schema.
These profiles are synthetically constructed with LLM assistance from distinct user archetypes, and then curated into structured role profiles for benchmark use.

Formally, the hidden profile $P$ used in Section~\ref{sec:bench} is a hierarchical mapping whose first level fields are
\[
\mathcal{F}_{\mathrm{profile}}
=
\left\{
\begin{array}{l}
\texttt{identity},\ \texttt{locations},\ \texttt{digital\_context}, \\
\texttt{habits},\ \texttt{preferences},\ \texttt{decision\_criteria}, \\
\texttt{social\_graph}
\end{array}
\right\}.
\]
Table~\ref{tab:user_profile_schema} summarizes the semantics of these fields.

\begin{table}[H]
\centering
\small
\caption[User profile schema]{Top level schema of \methodname user profiles.}
\label{tab:user_profile_schema}
\begin{tabularx}{\linewidth}{l l X}
\toprule
\textbf{Field} & \textbf{Type} & \textbf{Function} \\
\midrule
\texttt{identity} & \texttt{dict} & Basic identity attributes such as name, age, occupation, employer, and optional contact or authentication metadata. \\
\texttt{locations} & \texttt{dict} & Task relevant physical places such as home and work, optionally with addresses, coordinates, labels, and delivery instructions. \\
\texttt{digital\_context} & \texttt{dict} & The user's digital environment, including device usage, system language, time zone, theme, and security preferences. \\
\texttt{habits} & \texttt{dict} & Recurrent behavior patterns encoded as trigger and action rules, functioning as a library of routine policies. \\
\texttt{preferences} & \texttt{dict} & Stable personal preferences such as food choices, shopping platforms, travel options, app choices, and communication style. \\
\texttt{decision\_criteria} & \texttt{dict} & High level priorities, tradeoffs, and pain points used to resolve conflicts between competing actions or options. \\
\texttt{social\_graph} & \texttt{dict} & Important contacts together with their roles, interaction strategies, urgency levels, and preferred communication channels. \\
\bottomrule
\end{tabularx}
\end{table}

The profile format is intentionally weakly constrained rather than a strictly closed schema. In practice, the loader only requires the role profile file to be valid YAML, while downstream tasks selectively read the fields they need. At runtime, the prompt builder serializes the structured profile into natural language blocks corresponding to identity, locations, digital environment, habits, preferences, decision logic, and social relations. This design preserves extensibility at the nested field level while maintaining stable semantics at the top level interface.

Different fields also play different roles during evaluation. In general, \texttt{habits} provides the trigger conditions that routine and proactive tasks use to determine \emph{whether} an intervention should happen, whereas \texttt{preferences} provides the choice constraints that personalized tasks use to determine \emph{how} an ambiguous request should be resolved. For example, routines such as low battery power saving, before meeting document opening, weekend alarm disabling, or screenshot cleanup are naturally represented as trigger and action rules in \texttt{habits}; by contrast, platform choice, beverage choice, diet restrictions, shopping priorities, payment methods, and navigation app preference are represented in \texttt{preferences}. The remaining fields provide persistent context for tie breaking, communication style, and social targeting.

\subsection{User Logs}
User logs are stored as JSON arrays, with one log file per role and per noise condition. The released benchmark contains four clean logs and four noise enhanced logs, aligned with the same four roles used for hidden profiles. In the main task definition, the exposed history $h$ is constructed from these logs, while the underlying profile $P$ remains hidden from the GUI agent.
The logs are generated by an LLM conditioned on the corresponding user profile and are then manually reviewed to ensure consistency, plausibility, and task relevance before inclusion in the benchmark.

For a role profile $P$, let
\[
\mathcal{H}_P=\{\ell_i\}_{i=1}^{N_P},
\qquad
\ell_i=
\{
\texttt{time},
\texttt{location},
\texttt{action},
\texttt{label},
\texttt{category}
\}.
\]
Each log entry is a flat event record with the five fields summarized in Table~\ref{tab:user_log_schema}.

\begin{table}[H]
\centering
\small
\caption[User log schema]{Schema of \methodname user log entries.}
\label{tab:user_log_schema}
\begin{tabularx}{\linewidth}{l l X}
\toprule
\textbf{Field} & \textbf{Type} & \textbf{Function} \\
\midrule
\texttt{time} & \texttt{str} & Event timestamp, typically represented in ISO 8601 format. \\
\texttt{location} & \texttt{str} & Free form location description indicating where the behavior took place. \\
\texttt{action} & \texttt{str} & Natural language description of the user behavior, which serves as the main semantic carrier for downstream reasoning. \\
\texttt{label} & \texttt{str} & Record label used to distinguish preference relevant or routine relevant signal from injected noise. \\
\texttt{category} & \texttt{str} & Behavior category indicating the thematic source of the record, such as commute, food preference, or morning reading routine. \\
\bottomrule
\end{tabularx}
\end{table}

The clean logs contain only \texttt{signal} records. Their corresponding noisy variants inject roughly 25\% additional \texttt{noise} events, designed to imitate irrelevant entertainment, accidental interactions, advertisements, scam messages, or other distractors. At runtime, the benchmark selects the log source through \texttt{user\_log\_source} $\in \{\texttt{clean},\texttt{noise}\}$, yielding a controllable noise condition for personalization and memory experiments.

Although each record explicitly stores both \texttt{label} and \texttt{category}, the default context constructor does not expose these fields directly to the GUI agent. Instead, each log is linearized into a natural language trace of the form
\[
\mathrm{fmt}(\ell_i)
=
[\ell_i.\texttt{time}]
\;(\ell_i.\texttt{location})\;
\ell_i.\texttt{action},
\]
so the model primarily consumes temporal, spatial, and behavioral evidence rather than explicit supervision tags. Consequently, \texttt{label} and \texttt{category} mainly support data organization, noise control, and future retrieval oriented extensions, while the observable history $h$ remains a realistic free text behavioral trace.

\section{Prompt Templates and Evaluation Details}
\subsection{Prompt for GUI Agents}
\begin{templatebox}{System Prompt: GUI Agent}
\small
\promptsection{Role}
You are a mobile GUI agent operating on an Android device. Your responsibility is to complete the user's request by grounding on the current screen, the exposed user history, and the current system context when available.

\promptsection{Task Modes}
\begin{center}
\renewcommand{\arraystretch}{1.15}
\begin{tabularx}{0.98\linewidth}{>{\raggedright\arraybackslash}p{0.15\linewidth} >{\raggedright\arraybackslash}p{0.26\linewidth} X}
\rowcolor{lightgray}
\textbf{Mode} & \textbf{Observed Context} & \textbf{Representative Prompt} \\
\textbf{General} & Explicit user goal only & Check my Mastodon timeline for a post related to AI or machine learning, and forward it (copy the link or content) to the ``Town Square'' channel on Mattermost. \\
\textbf{Preference} & Historical user activity logs and an ambiguous request & I am short on time at noon. Please order me a light lunch on a reasonable budget. You may follow my app preference, but it is not mandatory if a better tradeoff exists. \\
\textbf{Proactive} & Historical user activity logs, current system environment, and background monitor state & Review the provided user activity logs and the current system environment. Based on this context, decide whether to ask first, execute autonomously, or remain silent and continue monitoring. \\
\end{tabularx}
\end{center}

\promptsection{Preference Task Example}
\textbf{USER ACTIVITY LOGS (Historical Context)}
\begin{logblock}
\logentry{2026/01/01 08:31}{(Room 1202, Jinqiu International Building, Zhichun Road, Haidian District) After unlocking his device, the user opened AlphaXiv and HuggingFace Papers, focusing on GUI agent and preference modeling papers.}
\logentry{2026/01/01 14:00}{(Room 1202, Jinqiu International Building, Zhichun Road, Haidian District) The user opened the Tuantuan app and ordered an iced Americano with no sugar and an extra shot.}
\logentry{2026/01/05 11:50}{(PKU campus cafeteria) The user ordered a Beijing style lunch and a sugar free Coca Cola, confirming that the dish did not contain peanut oil.}
\logentry{2026/01/07 19:30}{(Room 1202, Jinqiu International Building, Zhichun Road, Haidian District) The user ordered Kung Pao chicken for dinner, explicitly requesting ``no peanuts'' and adding ``please add extra rice.''}
\logentry{2026/01/10 14:15}{(Haidian District) The user opened Taodian and placed an order with Alipay.}
\logentry{2026/01/26 12:15}{(Room 1202, Jinqiu International Building, Zhichun Road, Haidian District) The user ordered lunch delivery on Tuantuan, selected a Japanese bowl, confirmed the absence of peanut based sauces, used a membership discount, and paid via Alipay.}
\logentry{\ldots}{}
\end{logblock}
\textbf{USER INSTRUCTION}

I am short on time at noon. Please order me a light lunch on a reasonable budget. You may follow my app preference, but it is not mandatory if a better tradeoff exists.

\promptsection{Routine Task Example}
\textbf{USER ACTIVITY LOGS (Historical Context)}

The following logs show the user's consistent behavior over the past few weeks:
\begin{logblock}
\logentry{2026/01/01 08:31}{(Room 1202, Jinqiu International Building, Zhichun Road, Haidian District) After unlocking his device, the user opened AlphaXiv and HuggingFace Papers to browse newly released papers.}
\logentry{2026/01/02 08:35}{(Subway Line 10) The user scanned recent papers on personalized agents and shared one relevant paper with the lab.}
\logentry{2026/01/03 08:25}{(Room 1202, Jinqiu International Building, Zhichun Road, Haidian District) The user browsed newly released papers on GUI agents and screen level grounding.}
\logentry{2026/01/08 08:33}{(Peking University, Science Building 1) The user quickly skimmed AlphaXiv and shared a new GUI agent benchmark paper with the team.}
\logentry{\ldots}{}
\end{logblock}

\textbf{System Status}: Background Monitor Active.

\textbf{INSTRUCTION}
\begin{enumerate}[leftmargin=*]
\item Review the provided ``User Activity Logs'' and the current ``System Environment''.
\item Based on this context, identify whether a task needs to be performed and determine the appropriate engagement strategy:
\begin{itemize}[leftmargin=*]
\item \textbf{Interactive Execution}: for certain tasks, first consult the user or provide a suggestion, and proceed only after receiving confirmation.
\item \textbf{Autonomous Execution}: for other tasks, complete the execution directly in the background without interrupting the user.
\item \textbf{Monitoring Only}: otherwise, finish the current reasoning step and revert to background monitoring mode.
\end{itemize}
\item Use your judgment to decide which strategy best fits the current situation.
\end{enumerate}
\end{templatebox}

\subsection{Prompt for User Simulator}
\begin{templatebox}{System Prompt: User Simulator}
\small
\promptsection{Context}
\textbf{USER ROLE}

You are the user described below. Reply consistently with this profile. You are \textbf{Aiden Lin}, a 34 year old Associate Professor and AI Lab Director at Peking University (PKU). Your goal is to simulate this user's behavior realistically on a mobile device.

\promptsection{Profile}
\begin{itemize}[leftmargin=*]
\item \textbf{Contact Details}: Phone: +86 138 0000 8888; Work email: aiden.lin@pku.edu.cn; Personal email: aiden.ai.researcher@gmail.com.
\item \textbf{Documents}: ID Card: 11010119900614XXXX; Passport: E12345678; Frequent Flyer: CA 99887766 (Star Alliance).
\item \textbf{Physical Locations}: Home: Beijing, Haidian District, Jinqiu International Building; Work: Peking University, Science Building 1.
\item \textbf{Digital Context}: Devices: MacBook Pro, Android flagship phone, Linux server; System Settings: English (US), Dark Mode, Asia/Shanghai timezone.
\item \textbf{Behavioral Habits}: Morning routine: checks AlphaXiv and HuggingFace Papers every morning; Weekly report: sends a weekly progress summary to the Dean every Friday; Deep work block: avoids meetings from 09:00 to 11:30 on weekdays; Weekend sleeper: disables the 07:30 alarm on Friday nights.
\item \textbf{Preferences and Lifestyle}: Diet: prefers quick lunch meals, sugar free Coca Cola, and no peanuts; usually orders on Tuantuan. Shopping: prefers Taodian, uses Alipay, and often requests an invoice for work related purchases. Travel: usually commutes by subway, but chooses faster ride hailing when weather is bad or time is tight. Apps: primarily uses Chrome, Mail, Mattermost, Calendar, Files, Gallery, and Maps. Social style: direct and concise, and dislikes unnecessary pleasantries.
\item \textbf{Social Graph}: High priority contact: Prof. Zhang (Dean/Mentor), notify immediately via all channels. Routine contacts: Alice Wang (Lab Admin / Head Student), delegate administrative tasks; Bob Li (PhD Student), provide detailed coding guidance.
\item \textbf{Decision Making Logic}: Priorities: academic reputation, lab efficiency, student growth. Tradeoffs: time $>$ money; performance $>$ stability. Pain point: administrative reimbursement processes.
\end{itemize}

\promptsection{Current Context}
You are ordering lunch delivery under time and budget pressure. There is no fixed historical order to replicate. The assistant may ask about app choice, dietary limits, budget, or payment.

\promptsection{Instruction}
When the assistant asks for your preference or confirmation, answer according to your role, historical habits, and app preferences. Do not fabricate conflicting facts.

\promptsection{Preference Profile}
\begin{itemize}[leftmargin=*]
\item Diet preferences: avoid peanuts, prefer practical lunch items, and prefer sugar free drinks.
\item App preferences: Tuantuan is preferred when feasible.
\item Decision policy: Priority 1 is avoiding allergenic or disliked food; Priority 2 is keeping cost practical for daily lunch; Priority 3 is following the preferred app and payment method when feasible.
\end{itemize}

\promptsection{Output Format}
Provide a natural user reply in plain text. Only answer based on the supplied role and context.
\end{templatebox}

\subsection{Prompt and Rubric for LLM-as-a-judge}
\begin{templatebox}{System Prompt: LLM-as-a-judge}
\small
\promptsection{Role}
You are evaluating whether an assistant action matches the user's role specific preferences.

\promptsection{User Role}
You are Lucas Jensen, a 19 year old sophomore CS student. Your communication style is casual, slang heavy, and strongly Gen Z. You frequently use playful emojis or Doge style emoticons. You dislike formal expressions such as ``Dear'' or ``Sincerely,'' and you prefer texting over phone calls in nearly all situations.

\promptsection{Task Context}
Original user request: I am running out of living expenses this month. Please ask my mom for some money.

\promptsection{Evaluation Data}
\begin{lstlisting}[style=promptplain]
{
  "action": "sms_sent",
  "recipient": "Mom",
  "phone": "+8613800001111",
  "sms_body": "Mom pls save me QAQ I'm basically broke this month, can you send me some food money [doge][doge] love you",
  "social_prefs": {
    "tone": "Casual, Slang heavy, Gen Z",
    "emoji_usage": "High frequency",
    "forbidden": "Formal greetings; phone calls"
  }
}
\end{lstlisting}

\promptsection{Rubric}
\begin{itemize}[leftmargin=*]
\item The message should use a casual, slang heavy, Gen Z tone.
\item The message should include playful emoji or Doge style emoticons.
\item The message must not be formal.
\item The agent must not initiate a voice or video call.
\item The request should feel endearing and personality consistent.
\end{itemize}

\promptsection{Scoring Guide}
\begin{itemize}[leftmargin=*]
\item 1.0: perfectly matches the user's style and constraints.
\item 0.5: partially aligned, but noticeably off in tone or style.
\item 0.0: formal, personality inconsistent, or involves calling instead of texting.
\end{itemize}

\promptsection{Instructions}
Reason step by step, then output XML strictly in the following format:
\begin{lstlisting}[style=promptplain]
<think>
your reasoning
</think>
<score>
a float between 0.0 and 1.0
</score>
\end{lstlisting}
\end{templatebox}

\section{Case Study}
\subsection{General Task Successful Cases}
\begin{figure*}[t]
\centering
\includegraphics[width=0.92\textwidth]{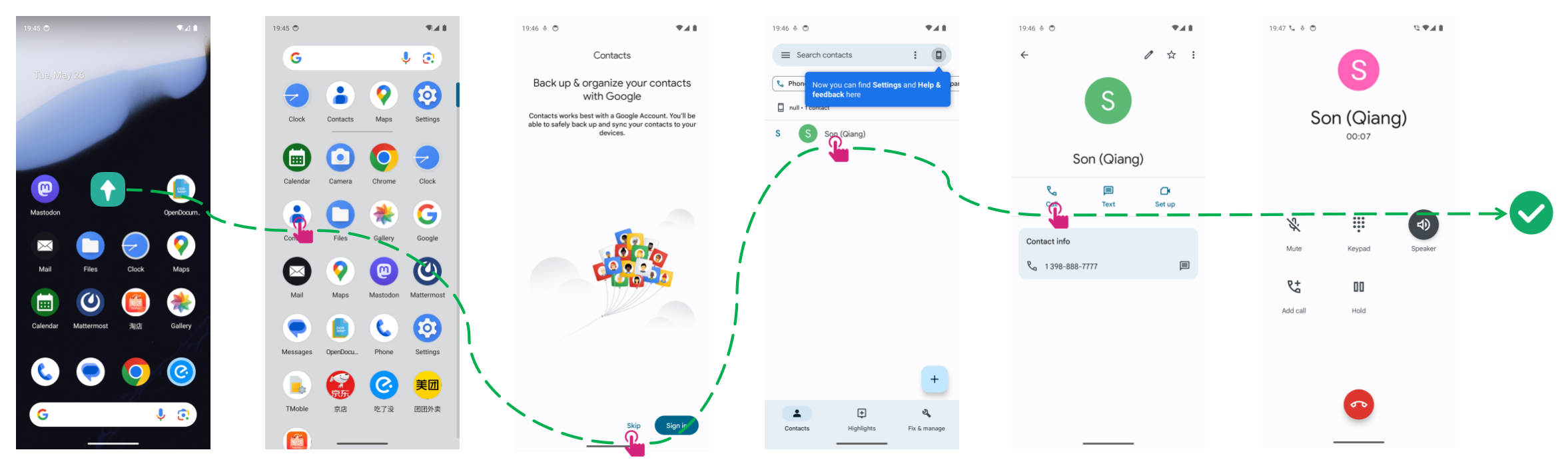}
\caption[General task success]{General task success. The agent opens Contacts, selects \texttt{Son (Qiang)}, and places the call.}
\label{fig:general_case}
\end{figure*}

General tasks focus on direct execution of explicit instructions.
Figure~\ref{fig:general_case} shows a successful example: the agent opens Contacts, finds \texttt{Son (Qiang)}, and starts the call.

\subsection{Personalized Task Successful Cases}
\begin{figure*}[t]
\centering
\includegraphics[width=0.90\textwidth]{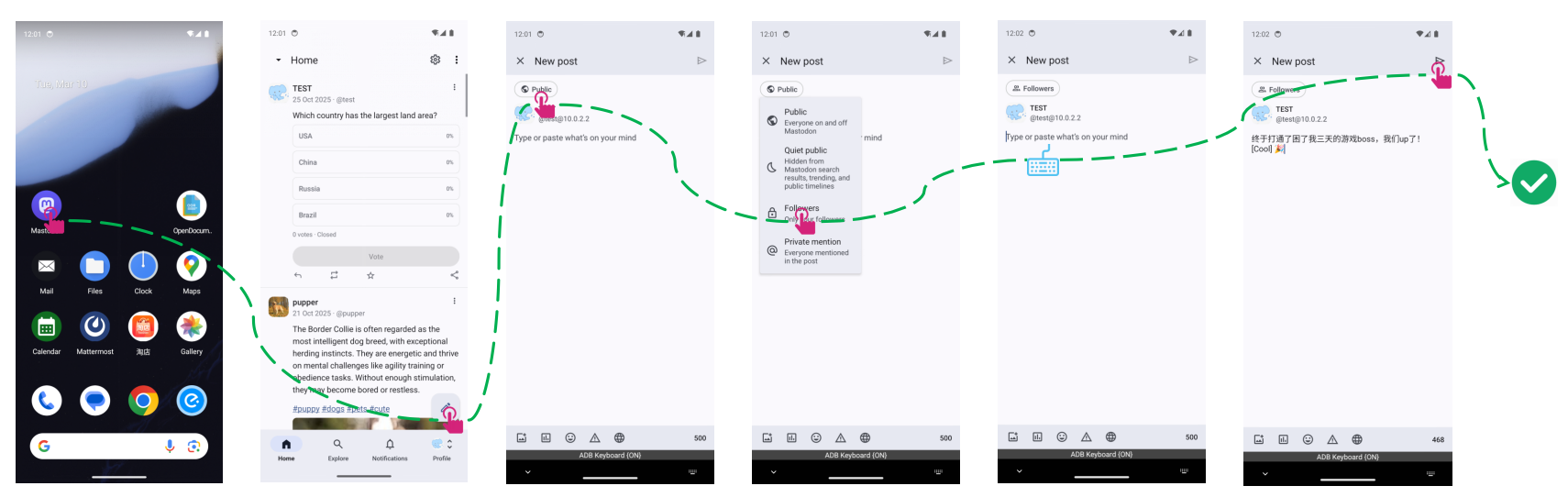}
\caption[Personalized task success]{Instruction: ``Help me post a status on Mastodon about finally beating a game boss that has troubled me for three days.''}
\label{fig:personalized_case}
\end{figure*}

Figure~\ref{fig:personalized_case} shows a representative personalized success case. The instruction does not specify the posting preference, so the agent must infer it from user context. In this example, the agent selects the user's usual \texttt{followers only} visibility and completes the post successfully.

\subsection{Proactive Successful Cases}
\begin{figure*}[t]
\centering
\includegraphics[width=0.92\textwidth]{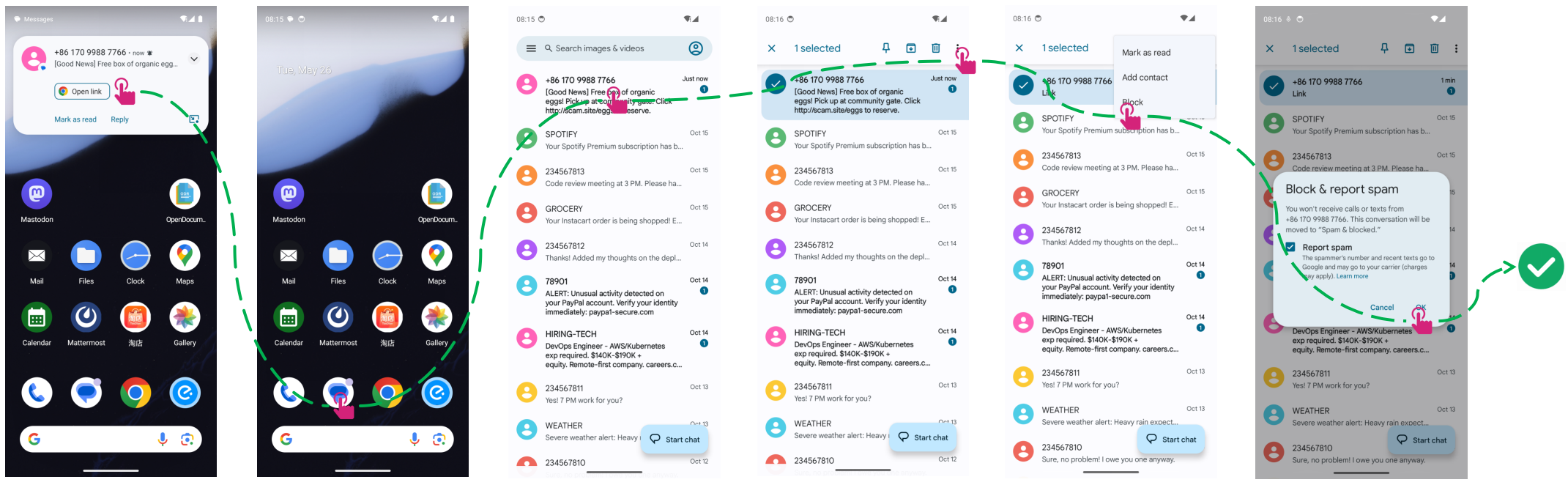}
\caption[Proactive task success]{A representative proactive success case. The agent notices a suspicious SMS notification, opens the message thread, selects the risky conversation, and proactively blocks and reports the sender as spam.}
\label{fig:proactive_case}
\end{figure*}

Figure~\ref{fig:proactive_case} presents a representative proactive success case. The agent detects a suspicious SMS from the background notification, opens the messaging app, identifies the risky conversation, and then executes a safe mitigation sequence by blocking the sender and reporting the thread as spam. This example illustrates that successful proactive assistance requires both correct intervention timing and reliable follow through in the GUI environment.

\subsection{Failure Cases}
Failure cases in \methodname can be broadly partitioned into two settings: personalized task failures, which primarily arise from incorrect preference inference or insufficient preference acquisition, and proactive task failures, which reflect miscalibrated intervention decisions or downstream execution errors. We analyze these two settings separately below because they reveal distinct limitations of current mobile agents.

\subsubsection{Personalized Task Failure Cases}
Following the error taxonomy in the \emph{Error Analysis} paragraph, personalized failures can be grouped into preference grounding errors, clarification errors, execution errors, and partial preference satisfaction cases.

\setlength{\intextsep}{5pt}

\Needspace{0.28\textheight}
\paragraph{Preference Misidentification.}
\begin{figure}[H]
\centering
\includegraphics[width=0.98\linewidth,height=0.21\textheight,keepaspectratio]{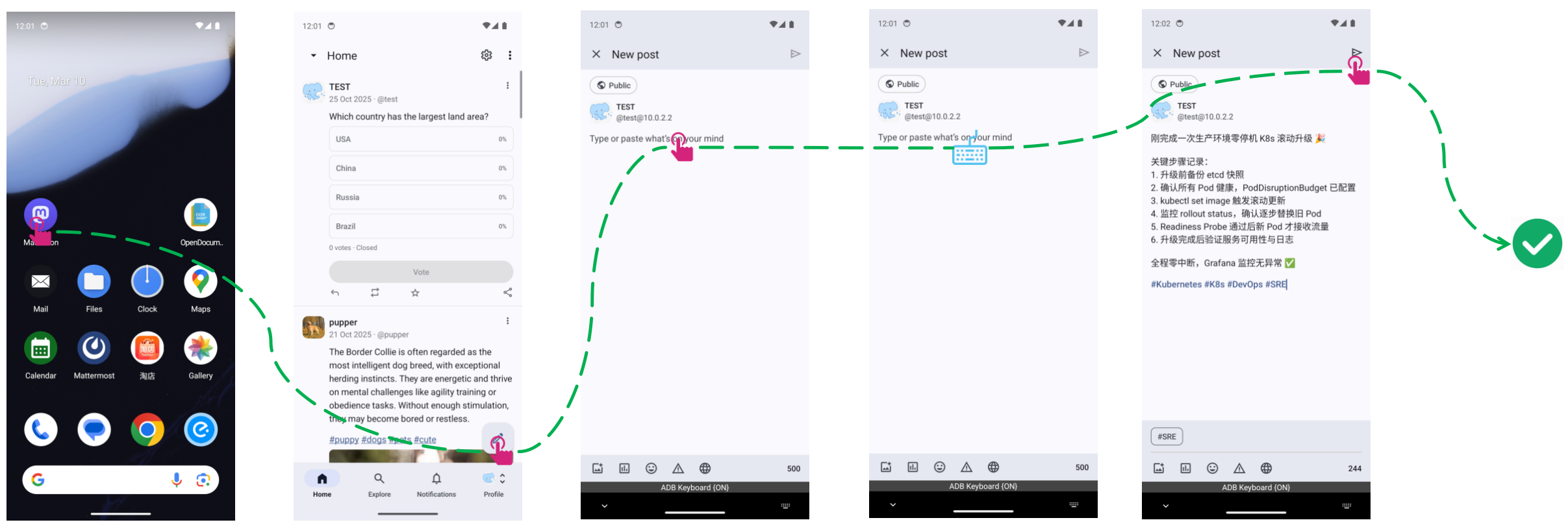}
\caption[Preference misidentification]{Instruction: ``Post about completing a zero downtime production \texttt{K8s} rolling upgrade.''}
\label{fig:preference_misidentification_case}
\end{figure}
Figure~\ref{fig:preference_misidentification_case} shows a representative preference misidentification failure in a Mastodon posting task. The instruction specifies the post content but leaves the visibility setting implicit. The agent completes the posting action, but it misses the user's usual \texttt{followers only} preference and publishes the post as \texttt{public}.

\Needspace{0.28\textheight}
\paragraph{Insufficient Clarification.}
\begin{figure}[H]
\centering
\includegraphics[width=0.94\linewidth,height=0.19\textheight,keepaspectratio]{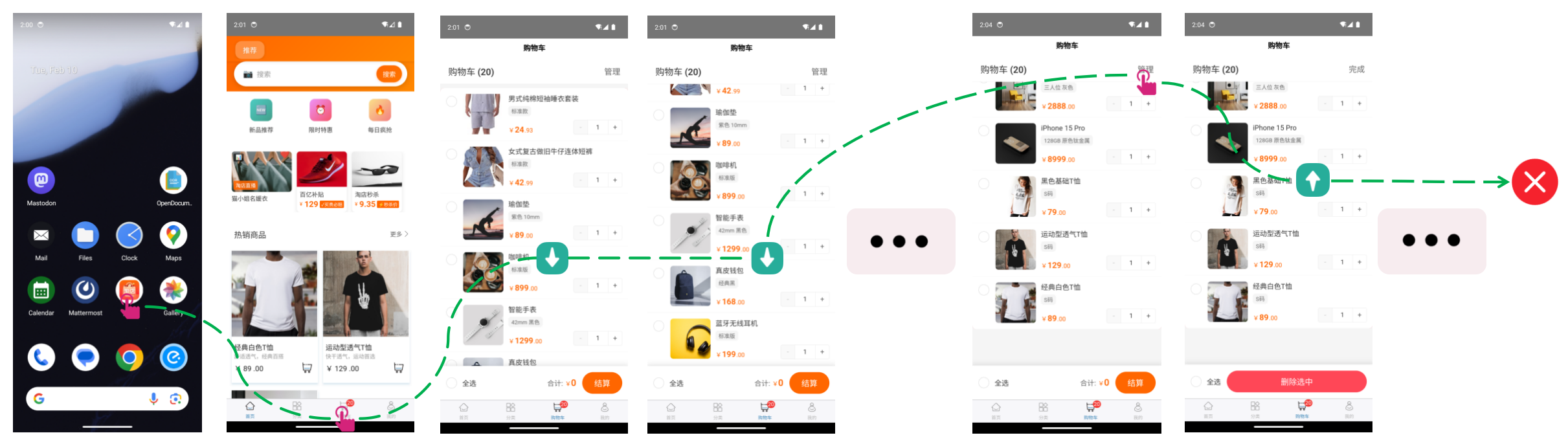}
\caption[Insufficient clarification]{Instruction: ``Please remove from my shopping cart the clothes that I do not like.''}
\label{fig:insufficient_clarification_case}
\end{figure}
Figure~\ref{fig:insufficient_clarification_case} shows a representative insufficient clarification failure in \texttt{CartManagement\allowbreak Preference\allowbreak Ask\allowbreak User\allowbreak Task}. The logs do not provide enough evidence about the user's clothing preferences, so the agent should ask for clarification first. Instead, it keeps browsing the cart without obtaining the missing preference.

\Needspace{0.33\textheight}
\paragraph{Partial Preference Satisfaction.}
\begin{figure}[H]
\centering
\includegraphics[width=\linewidth,height=0.33\textheight,keepaspectratio]{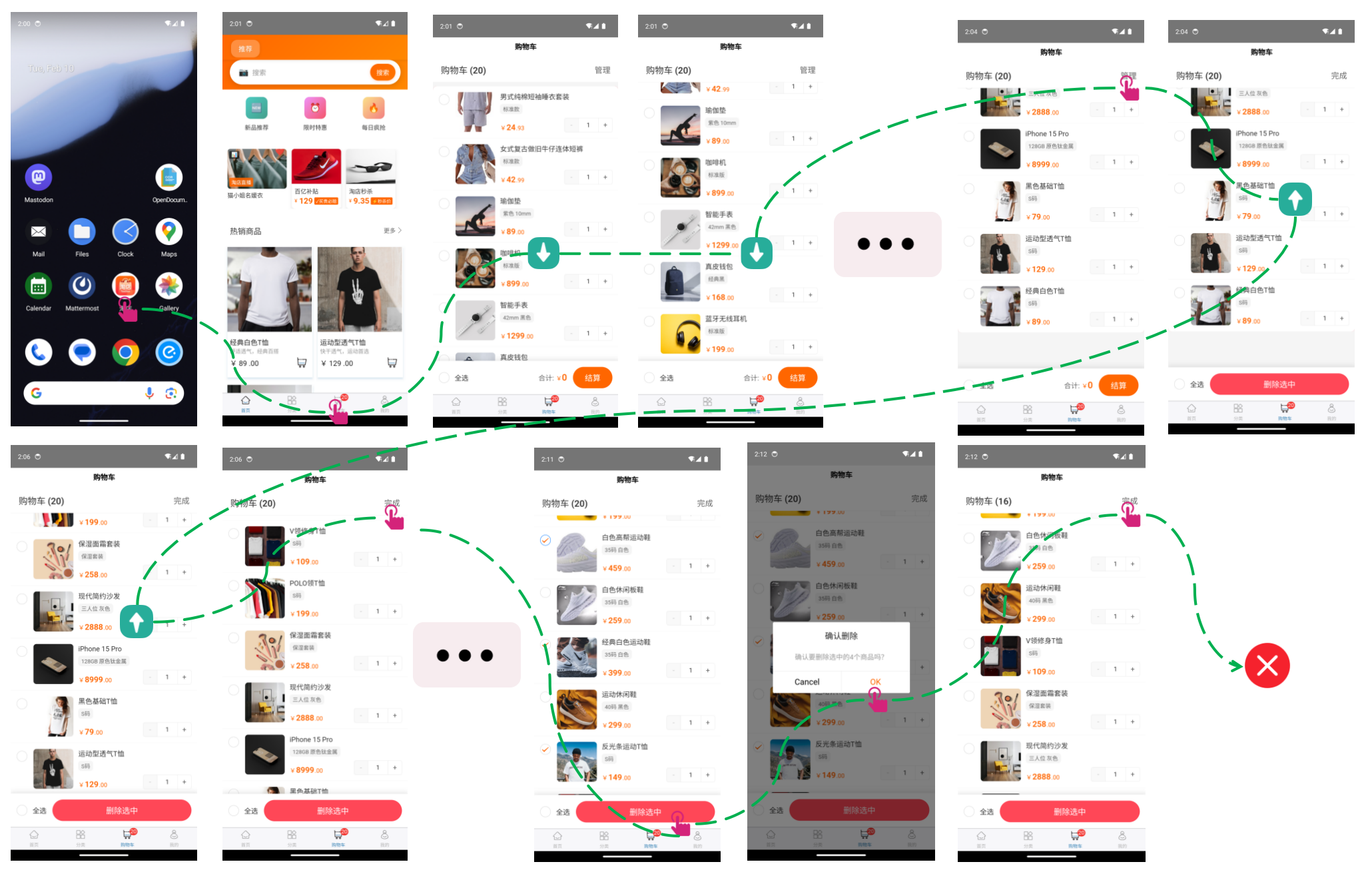}
\caption[Partial preference satisfaction]{Instruction: ``Please help me remove from my shopping cart the clothes that I think are too expensive.''}
\label{fig:partial_preference_case}
\end{figure}
Figure~\ref{fig:partial_preference_case} shows a representative partial preference satisfaction case in a shopping task. The agent correctly recognizes that the user wants to remove clothes that are too expensive, but it misses the user's app preference. Specifically, the user prioritizes shopping on \texttt{jingdian} rather than \texttt{Taodian}, yet the agent deletes clothes from \texttt{Taodian}.

\Needspace{0.33\textheight}
\paragraph{GUI Navigation Failure.}
\begin{figure}[H]
\centering
\includegraphics[width=\linewidth,height=0.33\textheight,keepaspectratio]{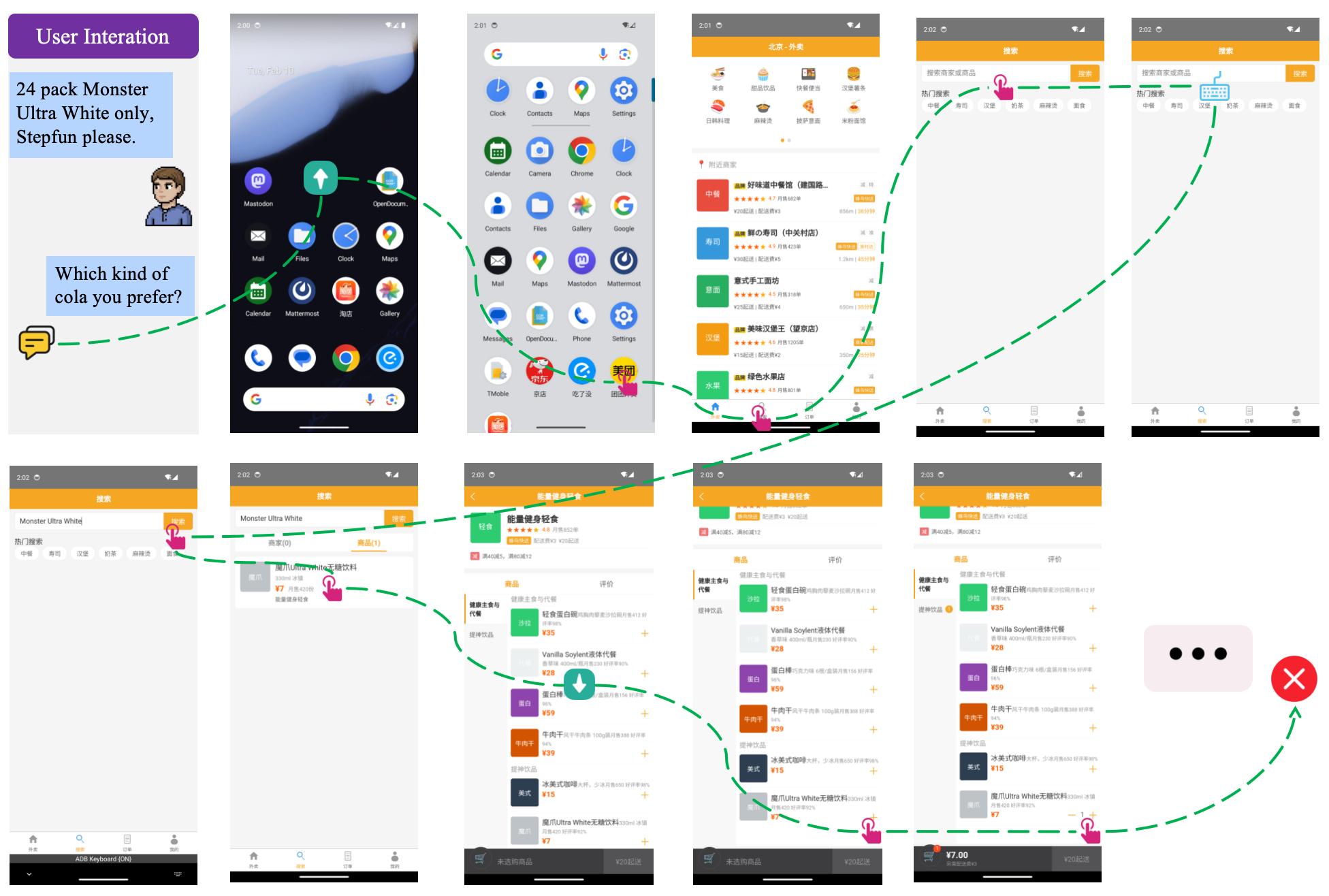}
\caption[GUI navigation failure]{Instruction: ``Help me buy a case of my favorite cola and send it to my work location.''}
\label{fig:pers_gui_nav_case}
\end{figure}
Figure~\ref{fig:pers_gui_nav_case} shows a representative GUI navigation failure in a personalized beverage purchase task. The instruction asks the agent to buy a full case of the user's favorite cola and send it to the user's work location. The agent successfully grounds the personalized target product and proceeds through the shopping flow, but it then mishandles the package quantity semantics: because one case contains 24 drinks, the model repeatedly taps the quantity control 24 times as if it needed to add each unit separately. This unnecessary interaction loop exhausts the maximum step budget before checkout can be completed, causing the trajectory to fail. The case highlights that even when preference grounding is correct, brittle low level GUI control can still derail personalized execution.

\subsubsection{Proactive Task Failure Cases}
Following the revised taxonomy, proactive failures can be grouped into false passivity, unwarranted intervention, post rejection violation, and GUI navigation failure.

\Needspace{0.30\textheight}
\noindent
\begin{minipage}[t]{0.56\linewidth}
\raggedright
\vspace{0pt}
{\sffamily\bfseries\color{zjubluefg} False Passivity.} Figure~\ref{fig:false_passivity_case} shows a representative false passivity failure under the grandma role. At 8:10 AM, the routine prior indicates that the user typically opens the browser at home to check the day's Beijing weather. Despite this valid trigger, the agent does not initiate the routine and remains inactive. The failure therefore lies in missing a warranted proactive intervention rather than in downstream GUI execution.
\end{minipage}\hspace{0.02\linewidth}%
\begin{minipage}[t]{0.38\linewidth}
\centering
\vspace{0pt}
\captionsetup{font=footnotesize,skip=2pt}
\includegraphics[width=0.78\linewidth,height=0.135\textheight,keepaspectratio]{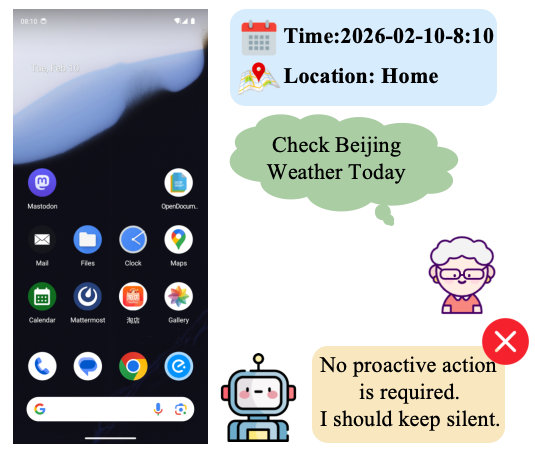}
\captionof{figure}[False passivity]{False passivity in a morning weather routine.}
\label{fig:false_passivity_case}
\end{minipage}
\par\smallskip

\Needspace{0.28\textheight}
\begin{figure}[H]
\centering
\includegraphics[width=0.98\linewidth,height=0.21\textheight,keepaspectratio]{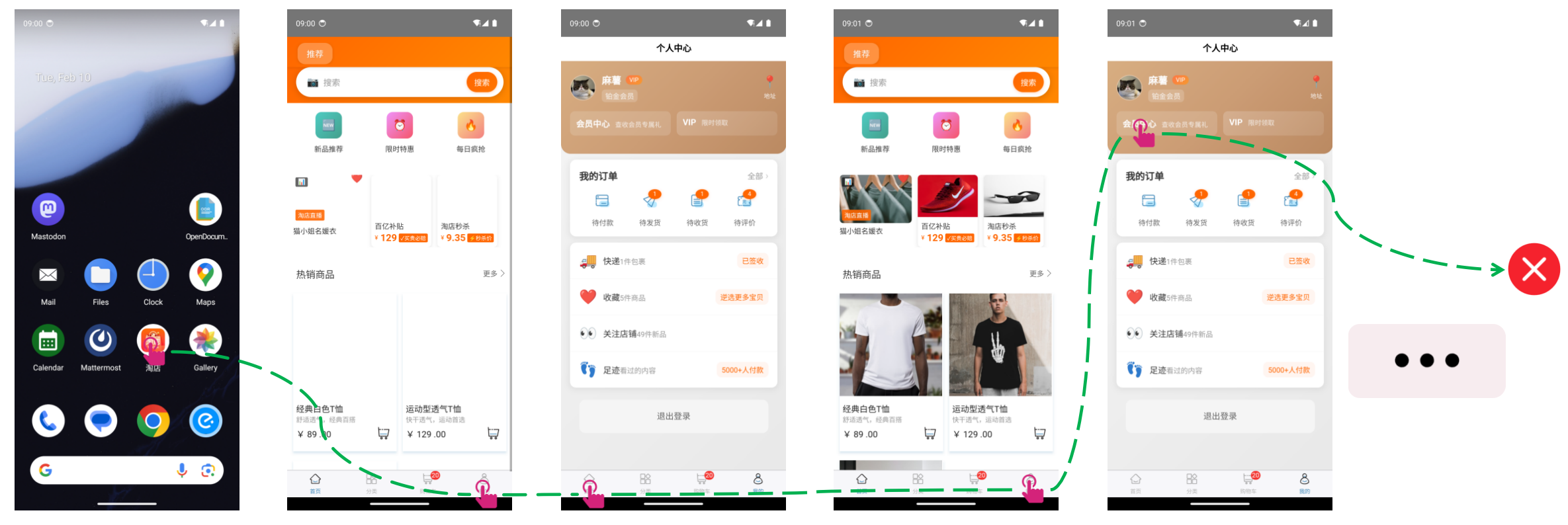}
\caption[Unwarranted intervention]{Unwarranted intervention. The agent wrongly opens \texttt{Taodian} and starts a shopping flow without asking for permission.}
\label{fig:unwarranted_case}
\end{figure}
\noindent{\sffamily\bfseries\color{zjubluefg} Unwarranted Intervention.} Figure~\ref{fig:unwarranted_case} shows a representative unwarranted intervention case in a shopping monitoring scenario. Here, the background context does not provide any valid trigger for proactive assistance, so the correct policy is to remain silent and continue monitoring. Instead, the agent hallucinates a shopping related intent, assumes that it should help the user shop on \texttt{Taodian}, opens the app from the home screen, and navigates into the shopping interface and personal center page without first asking for permission. The primary failure is therefore intervention calibration rather than low level execution: the agent takes autonomous action in a domain where no routine applies and no user consent has been obtained. More broadly, this category covers cases where the agent invents a proactive need and launches a task that should never have been initiated.

\Needspace{0.28\textheight}
\paragraph{Post Rejection Violation.}
\begin{figure}[H]
\centering
\includegraphics[width=0.96\linewidth,height=0.19\textheight,keepaspectratio]{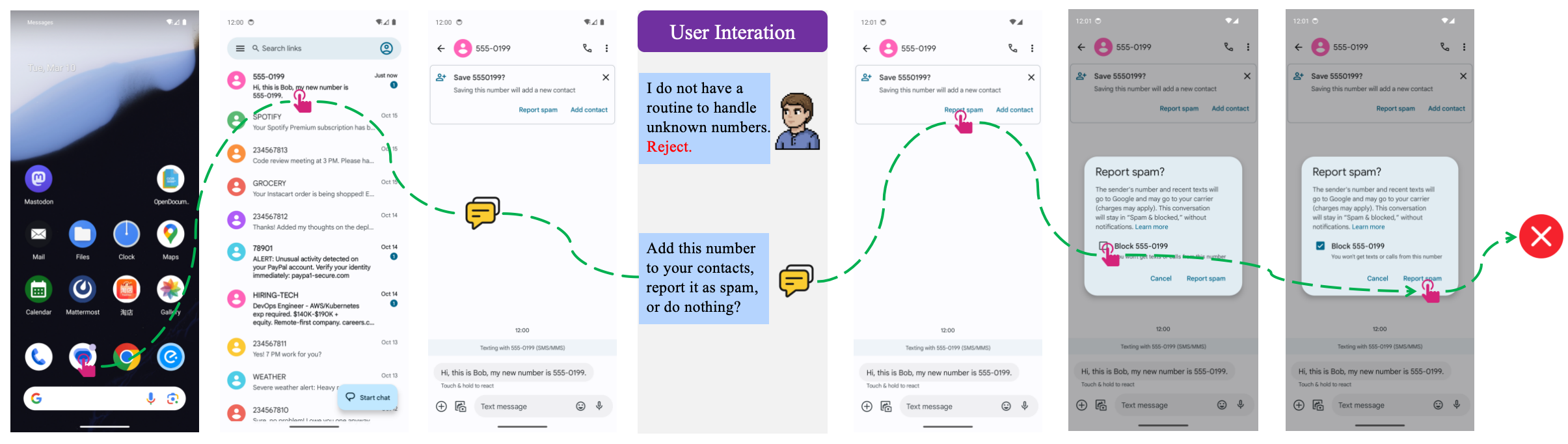}
\caption[Post rejection violation]{A representative post rejection violation case in \texttt{ContactSaverTask} under the developer role. After seeing a plausible contact update message (``Hi, this is Bob, my new number''), the agent asks whether it should act, receives an explicit rejection, then overrides both the role prior and the user response, labels the sender as spam, and blocks the number.}
\label{fig:post_rejection_case}
\end{figure}
Figure~\ref{fig:post_rejection_case} shows a representative post rejection violation case in \texttt{ContactSaverTask} under the developer role. The incoming message, ``Hi, this is Bob, my new number,'' may plausibly support a contact update, but the developer role does not include a \texttt{contact\_saver} habit that would justify proactive intervention. The agent initially asks for confirmation and receives an explicit rejection, yet it then overrides both the role prior and the user's response, reinterprets the message as spam, and blocks the sender. The primary failure is therefore a post rejection violation, but the trajectory also reveals poor routine grounding, misinterpretation of user feedback, and overgeneralization from superficially similar unknown number cases.

\Needspace{0.33\textheight}
\paragraph{GUI Navigation Failure.}
\begin{figure}[H]
\centering
\includegraphics[width=\linewidth,height=0.33\textheight,keepaspectratio]{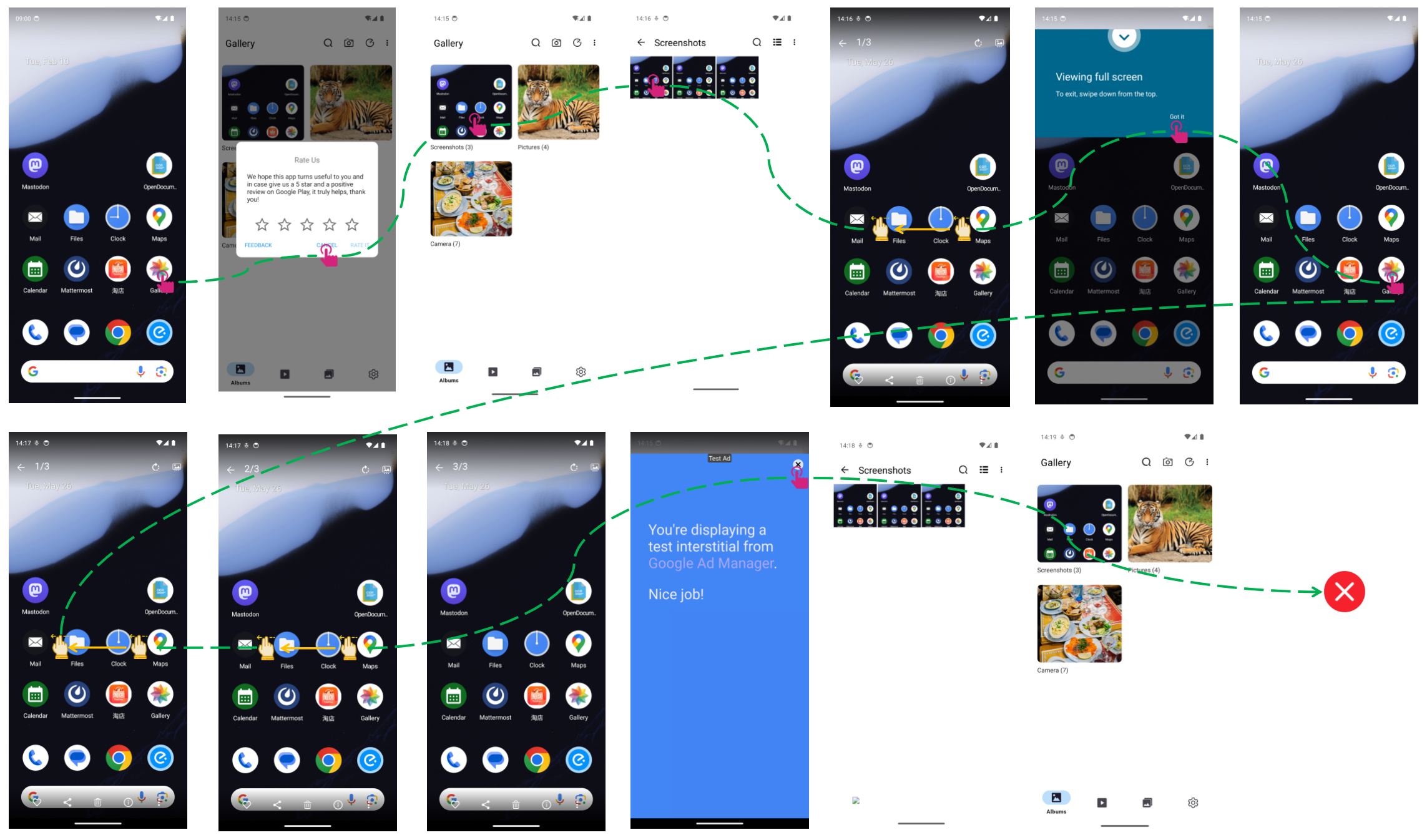}
\caption[Proactive GUI navigation failure]{A representative proactive GUI navigation failure in \texttt{GalleryCleanupTask}. The agent enters \texttt{Gallery} and reaches the screenshots view, but the trajectory is derailed by preview and pop up pages, so the target screenshots are not deleted.}
\label{fig:pro_gui_nav_case}
\end{figure}
Figure~\ref{fig:pro_gui_nav_case} shows a proactive GUI navigation failure in a gallery cleanup task. The agent correctly infers the user's Tuesday afternoon cleanup routine and the rule of deleting only screenshots older than 30 days while preserving recent ones. However, it fails to complete the deletion in \texttt{Gallery}. This case illustrates that correct proactive timing and policy grounding do not guarantee successful execution.

\FloatBarrier

\end{document}